%% file: main.tex
\documentclass{article} 
\usepackage{iclr2023_conference,times}

\input{math_commands.tex}

\usepackage{hyperref}
\usepackage{url}

\usepackage{tikz}
\usepackage{booktabs}
\usepackage{multirow}
\usepackage{todonotes}
\usepackage{listings}
\usepackage{censor}
\usepackage{pgfplots}
\DeclareUnicodeCharacter{2212}{−}
\usepgfplotslibrary{groupplots,dateplot}
\usetikzlibrary{patterns,shapes.arrows}
\pgfplotsset{compat=newest}
\usepackage{bbm}
\usepackage{caption, subcaption}

\newcommand{\ourMethodName}{Polite Teacher}

\title{\ourMethodName: Semi-Supervised Instance Segmentation with Mutual Learning and Pseudo-Label Thresholding}

\author{
    Dominik Filipiak$^{1,2,3}$\thanks{\url{dfilipiak@mimuw.edu.pl}},
    Andrzej Zapa\l{}a$^{3}$,
    Piotr Tempczyk$^{1,3,4,5}$,
    Anna Fensel$^{6,2}$,
    Marek Cygan$^{3}$ \\
    $^1$AI Clearing, Inc. \quad
    $^2$University of Innsbruck \quad
    $^3$University of Warsaw \quad
    $^4$\url{opium.sh} \\
    $^5$\url{deeptale.ai} \quad
    $^6$Wageningen University \& Research
}

\iclrfinalcopy 
\begin{document}

\maketitle

\begin{abstract}
    We present \ourMethodName, a simple yet effective method for the task of semi-supervised instance segmentation.
    The proposed architecture relies on the Teacher-Student mutual learning framework.
    To filter out noisy pseudo-labels, we use confidence thresholding for bounding boxes and mask scoring for masks.
    The approach has been tested with CenterMask, a single-stage anchor-free detector.
    Tested on the COCO 2017 \texttt{val} dataset, our architecture significantly (approx. +8 pp. in mask AP) outperforms the baseline at different supervision regimes.
    To the best of our knowledge, this is one of the first works tackling the problem of semi-supervised instance segmentation and the first one devoted to an anchor-free detector. 
    The code is available.
\end{abstract}

\input{1_intro}
\input{2_survey}
\input{3_method}
\input{4_results}
\input{5_summary}

\input{6_contributions}
\input{7_acknowledgements}
\input{8_reproducibility_statement}
\input{9_ethics_statement.tex}

\bibliography{bibliography}
\bibliographystyle{iclr2023_conference}


\end{document}

%% file: math_commands.tex
\usepackage{amsmath,amsfonts,bm}

\def\eqref#1{equation~\ref{#1}}

\def\1{\bm{1}}

\DeclareMathAlphabet{\mathsfit}{\encodingdefault}{\sfdefault}{m}{sl}
\SetMathAlphabet{\mathsfit}{bold}{\encodingdefault}{\sfdefault}{bx}{n}



%% file: 1_intro.tex
\section{Introduction}
\label{sec:1_introduction}

The advent of deep learning transformed computer vision pipelines both in academia and industry.
However, progress is often hindered, since deep learning models are expensive to train for several reasons.
Leaving the hardware and computational expenses aside, the vast share of costs often comes from providing the right amount of samples to learn from.
For a number of supervised problems in computer vision, it is relatively easy to obtain data.
However, labelling them is often the real source of expenses.
Semi-supervised learning methods are tailored to deal with the situation in which there are \emph{enough} data samples, but access to the labels is severely limited.

Semantic segmentation (sometimes called dense classification) is a classical computer vision task of assigning each pixel a category.
This enables clustering images into semantically coherent parts.
Object detection is concerned with the location and identification of semantic objects on images.
Instance segmentation combines these two -- it is concerned with locating and identifying entities with pixel-wise accuracy.
While an intense research activity can be observed in the areas of semi-supervised semantic segmentation \citep{tarvainen2017mean,chen2021semi} and object detection \cite{liu2021unbiased,liu2022unbiased}, very little attention has been devoted to semi-supervised instance segmentation methods in the computer vision community.

We propose \ourMethodName, a simple yet effective method for the task of semi-supervised instance segmentation.
The architecture is built on the Teacher-Student framework.
Its \emph{politeness} in the name is an acronym for \emph{pseudo-label thresholding}, which is concerned with filtering noisy pseudo-labels in detection and mask heads.
Our contribution is two-fold:
\begin{itemize}
    \item We present \ourMethodName~-- one of the first works devoted to semi-supervised instance segmentation and the first one devoted to modern anchor-free detectors. 
    Our approach uses mask scoring \citep{huang2019mask} for the pseudo-mask thresholding.
    \item The presented method significantly (approx. +8 pp. in mask AP) improves baseline performance with different supervision regimes on COCO 2017 and sets the new baseline for further comparison, becoming de facto the new state-of-the-art for this dataset.
\end{itemize}

The paper is organised as follows.
Section \ref{sec:2_survey} contains a comprehensive survey of related research.
In Section \ref{sec:3_method}, we present the details of our method -- \ourMethodName.
The results of the evaluation are discussed in Section \ref{sec:4_results}, along with the detailed analysis and ablation studies.
The paper is concluded with a short summary in Section \ref{sec:5_summary}.

%% file: 2_survey.tex
\section{Related research}
\label{sec:2_survey}

This section presents a comprehensive survey of areas adjacent to our work.
We present recent research in the area of instance segmentation.
Then, we summarise the recent progress in semi-supervised learning.
Finally, we combine these two areas and briefly discuss the body of knowledge for semi-supervised instance segmentation. 

\paragraph{Instance segmentation.}
Instance segmentation is a computer vision problem concerned with pixel-wise delineating instances of semantic classes.
As it can be perceived as a combination of object detection and semantic segmentation, the advances in instance segmentation are tightly coupled with the two aforementioned tasks (especially the former).
In recent years, two kinds of object detectors were popular: single- and two-staged.
A typical example of the latter category is Faster R-CNN \citep{ren2015faster}.
It consists of the backbone feature extractor (eg. ResNet) and two heads: the region proposals network (RPN) and the second one for final detections (RoI -- the region of interests head).
The proposal candidates are searched on a pre-defined set of anchors using RPN and they are later refined with the RoI head.
\cite{he2017mask} introduced Mask R-CNN, which added the mask head to Faster R-CNN to solve segmentation tasks on predicted bounding boxes.
Mask scoring \citep{huang2019mask} adds another head on top of that -- it regresses the IoU (intersection over union) score of the predicted masks to improve model robustness.
Single-stage detectors try to achieve the outcomes of the aforementioned architecture in a single pass.
This often results in higher speed at the expense of precision.
A notable examples of such detectors include the YOLO family \citep{redmon2016you} or RetinaNet \citep{lin2017focal}.
More recently, \cite{lee2020centermask} proposed CenterMask, an anchor-free instance segmentation framework targeted at real-time applications.
It is build on FCOS \citep{tian2020fcos}, which details are discussed in Section \ref{sec:3_method}.
The architecture of CenterMask2 introduces spatial attention-guided masks (SAG-Mask) along with backbone feature extractors tailored for instance segmentation.
Recently, there is a surge of research on architectures utilising the concept of self-attention (also called transformers).
\cite{dosovitskiy2020image} introduced the visual transformer (ViT), which successfully adapted self-attention to computer vision.
This seminal work has sparked research interest in transformers in the vision community.
For instance, DINO \citep{caron2021emerging} adapts the Teacher-Student paradigm and self-supervised learning for various vision tasks, such as object detection.
Recently, MaskDINO \citep{li2022mask} added mask prediction to DINO and topped several instance segmentation benchmarks.
However, while displaying exceptional performance, solutions purely based on ViT suffer from quadratic computational complexity, which hinders their adoption.

\paragraph{Semi-supervised learning.}
Semi-supervised learning techniques can be framed as a middle ground between supervised and unsupervised learning since data with and without labels participate in the learning process.
It is related to weakly supervised and self-supervised learning.
Some approaches to the problem of semi-supervised learning such as $\Gamma$ model \citep{rasmus2015semi}, $\Pi$ model or temporal ensembling \citep{laine2017temporal} use the notion of self-ensembling.
However, more modern ones are focused on the non-standard architecture during the training phase, often incorporating multiple subnetworks.
For instance, \cite{tarvainen2017mean} introduced Mean Teacher, which is a popular semi-supervised training framework.
It overcomes the limitations of Temporal Ensembling and $\Pi$ models.
Instead of using the standard gradient-based approach, the teacher is updated using the exponential moving average (EMA).
Unbiased Teacher \citep{liu2021unbiased} builds on top of the Mean Teacher framework -- it does add focal loss and confidence thresholding of pseudo labels.
Focal loss borrowed from the work of \cite{lin2017focal} helps with the class imbalance, whereas confidence bounding box thresholding reduces the influence of noisy pseudo-labels.
The recent Ubiased Teacher v2 \citep{liu2022unbiased} extends it to anchor-free detectors and tackles the issue of the pseudo-labelling on bounding box regression.
Besides the Teacher-Student paradigm, there are also other approaches.
Cross pseudo supervision for semantic segmentation \citep{chen2021semi} is an example of a consistency regularisation method.
Here, two networks are trained on the output of each other and are penalised for discrepancies in predictions.
Contrastive methods constitute another approach to semi-supervised learning.
For instance, regional contrast, abbreviated as ReCo \citep{liu2022reco} belongs to this category.
While using the Teacher-Student framework, this model introduces a dedicated loss function and utilises the semantic relationship between classes.

\paragraph{Semi-supervised instance segmentation.}
Contrary to object detection and semantic segmentation, instance segmentation in the semi-supervised setting received little attention among scholars so far.
Concurrently to our work, \cite{wang2022noisy} presented Noisy Boundaries (NB).
This framework also uses the Teacher-Student paradigm and introduces different bounding box thresholds per category, drawing from the work of \cite{radosavovic2018data}.
The NB architecture has also two special features: the noise-tolerant mask head and boundary preserving re-weighting.
While the noise-tolerant head works with low-level resolution features to suppress the noise on mask boundaries, the boundary preserving map is focused on highlighting the boundary region for the segmentation part.
At the time of writing this publication, the problem of semi-supervised instance segmentation with anchor-free detectors has never been tackled in the literature.

%% file: 3_method.tex
\section{\ourMethodName}
\label{sec:3_method}

This section is devoted to the introduction of \ourMethodName.
First, we formulate the problem we are solving -- semi-supervised instance segmentation.
Then, we introduce the architecture of our solution -- used detectors, the Teacher-Student learning paradigm, and pseudo-label thresholding.
The section is concluded with a detailed description of the used loss function.

\paragraph{Problem formulation.}
We consider the problem of \emph{semi-supervised instance segmentation}.
Instance segmentation is a computer vision task which combines object detection and semantic segmentation.
Semi-supervised setting means that only part of the data available during the training phase is labelled.
More formally, we consider training dataset $\mathcal{D}$ consisting of a set of $N_{\texttt{sup}}$ labelled ($\mathcal{D}^{\text{sup}}=\{ \mathbf{x}_i, \mathbf{y}_i \}_{i=1}^{N_{\texttt{sup}}}$) and $N_{\texttt{unsup}}$ unlabelled ($\mathcal{D}^{\text{unsup}} = \{ \mathbf{x}_i\}_{i=1}^{N_{\texttt{unsup}}}$) images.
Here, $\mathbf{x}_i$ and $\mathbf{y}_i$ stand for images and their labels (instances categories along with their bounding boxes and masks) respectively.
Typically, $N_{\texttt{unsup}} >> N_{\texttt{sup}}$.
In this work, we assume that $\mathcal{D}^{\text{sup}}$ and $\mathcal{D}^{\text{unsup}}$ come from the same distribution.

\subsection{Architecture}

The architecture of \ourMethodName~depends on several components.
The first one is the detector, which is used twice due to the Teacher-Student paradigm.
We use CenterMask \citep{lee2020centermask}.
This is a single-stage anchor-free detector, which has a relatively simple architecture and therefore it is easy to tune.
Two such networks are then framed in the Teacher-Student paradigm to handle both labelled and unlabelled data.
Finally, two-fold pseudo-label thresholding takes place to remove noisy ones.
The first one uses bounding box uncertainty, and the second one rejects masks with an estimated low IoU score.

\paragraph{Detector.}
To properly present CenterMask, FCOS should be discussed first.
\cite{tian2020fcos} introduced Fully Convolutional One-Stage Object Detector (abbreviated as FCOS), an anchor-free object detector.
In general, one-stage detectors due to the lack of the proposal generation phase have fewer hyper-parameters to tune and therefore there are easier to train. 
Being anchor-free means eliminating pre-defined anchors, which diminishes the computational burden related to calculating IoU scores.
FCOS frames detection as a per-pixel prediction task, which resembles semantic segmentation.
Three loss components are subject to optimisation: classification, regression, and centre-ness.
While classification works similarly to other detectors, the regression targets are quite different.
Instead of predicting bounding box corners (like in e.g. Faster R-CNN), the four regressed values are $l$ (the distance from the centre of a bounding box to its left border), $t$ (top), $r$ (right), $b$ (bottom).
Finally, the centre-ness denotes the centre of a given bounding box.
Ground-truth centre-ness for $(l^{*}, t^{*}, r^{*}, b^{*})$ is defined as follows:
\begin{equation}
    \text{centreness}^{*} = \sqrt{\frac{\min (l^{*}, r^{*})}{\max (l^{*}, r^{*})} \times \frac{\min (t^{*}, b^{*})}{\max (t^{*}, b^{*})}}.
\end{equation}
Intuitively, this approach promotes bounding boxes which are located at the centre of a given object.

\cite{lee2020centermask} introduced CenterMask, which extends FCOS for the task of instance segmentation.
This is done similarly to how Faster R-CNN \citep{ren2015faster} is extended by Mask R-CNN \citep{he2017mask}.
However, there are some differences.
For instance, the RoI assignment function is redefined due to the different levels of the feature pyramid (FPN) which are used.
Instead of the mask head from Mask R-CNN, CenterMask utilises the spatial attention-guided mask (abbreviated as SAG-Mask).
For $\mathbf{x}$, which here mean features extracted from RoI align, the attention-guided feature map is calculated as follows:
\begin{equation}
    \mathbf{x}_{\text{sag}} = \sigma\left( \text{conv}_{3 \times 3} \left( \text{concat}\left( P_{\text{max}}, P_{\text{avg}}\right)\right) \right) \odot \mathbf{x},
\end{equation}
where $\sigma$ denotes sigmoid function, $\text{conv}_{3 \times 3}$ is convolutional layer with $3 \times 3$ filter, $P_{\text{max}}$ and $P_{\text{avg}}$ are the results of max and average pooling, and $\text{concat}$ stands for the concatenation. 

\paragraph{Teacher-Student Learning.}
We adopt a 2-step training procedure.
In the first step, the model is trained using only labelled data ($\mathcal{D}^{\text{sup}}$), which makes this part a standard supervised instance segmentation.
Instead of using a fixed number of batches for this step -- as \emph{burn-in stage} in Unbiased Teacher \citep{liu2021unbiased} -- we rather train it as long as it converges in terms of mask AP and take the best model $\theta$ to ensure the highest results.
Naturally, this step is expected to take longer with a higher number of supervised examples.
In the second step, mutual Teacher-Student learning with pseudo-labels takes part.
The best model from the first step is used and copied to be used as student and teacher models ($\theta_s \leftarrow \theta$, $\theta_t \leftarrow \theta$).
The model can be trained with the burn-in stage as well.

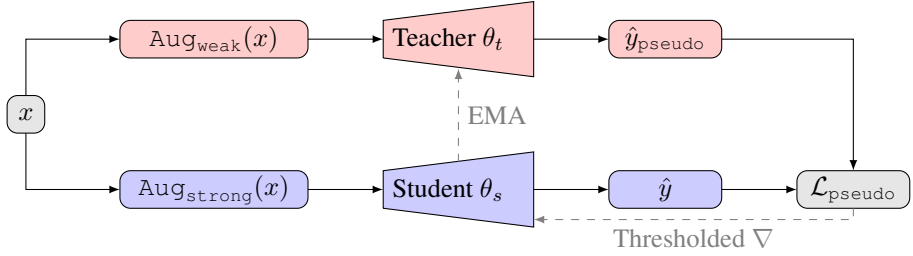
\begin{figure}
    \input{figures/architecture.tikz}
    \caption{Architecture of \ourMethodName~(the unsupervised part).}
    \label{fig:architecture}
\end{figure}

Teacher and student models receive the same input data -- they are augmented differently, though.
The teacher receives moderately augmented images (\emph{weak} augmentations -- we use random flipping), whereas the student consumes visibly perturbed images (\emph{strong} augmentations -- same as weak, plus colour jitter, random grayscale, gaussian blur, and random patch erasing).
During the training, the predictions from the teacher model serve as pseudo-labels (bounding boxes with their classes and masks) for the student.
The teacher is updated using the exponential moving average -- see equations \ref{eq:optim_student} and \ref{eq:optim_teacher} in the next subsection.
Figure \ref{fig:architecture} illustrates the unsupervised part of the process (the second step).

\paragraph{Pseudo-Label Thresholding.}
As the teacher is used to generate pseudo-labels $\mathbf{\hat{y}}$ in the semi-supervision regime, they can be noisy -- especially with a high share of unsupervised data.
Therefore, \ourMethodName~uses two-step pseudo-label thresholding: one is concerned with bounding boxes, whereas the second one refines the masks.
Similarly to STAC \cite{sohn2020simple} and Unbiased Teacher \citep{liu2021unbiased}, we introduce a bounding box confidence threshold -- $\tau_{\text{cls}}$.
Bounding boxes with a classification score smaller than $\tau_{\text{cls}}$ are discarded and not used further in the training.
The sigmoid output of the classification is treated here as confidence.
Inspired by the work of \cite{huang2019mask}, we also use a mask-scoring mechanism.
It regresses the IoU values of the generated masks and improves instance segmentation performance due to the prioritisation of more accurate masks.
While not directly designed for the task of semi-supervised learning, the output of this block can be used for filtering noisy pseudo-masks in a straight-forward manner.
That is, only masks satisfying $\mathbf{\hat{y}_{\text{IoU}}} > \tau_{\text{IoU}}$ are used in the unsupervised learning stage.
The other ones are considered uncertain and receive zero gradients.

\subsection{Optimisation}

The overall batch-wise loss $\mathcal{L}$ for supervised $\left\{ \left( \mathbf{x}_i, \mathbf{y}_i \right) \right\}_{i=1}^{B_{\text{sup}}}$ and unsupervised $\left\{ \left( \mathbf{x}_j, \mathbf{\hat{y}}_j \right) \right\}_{j=1}^{B_{\text{unsup}}}$ examples in a batch is computed as follows:
\begin{equation}
    \mathcal{L} = \sum_{i}^{B_{\text{sup}}} \mathcal{L}^{\text{sup}} \left( \mathbf{x}_i, \mathbf{y}_i \right) + \lambda \sum_{j}^{B_{\text{unsup}}} \mathcal{L}^{\text{unsup}} \left( \mathbf{x}_j, \mathbf{\hat{y}}_j \right),
    \label{eq:loss_main}
\end{equation}
where $\mathcal{L}^{\text{sup}}$ is the loss of the supervised part, and $\mathcal{L}^{\text{unsup}}$ is the loss of the unsupervised part.
The unsupervised part is scaled by $\lambda$.
The supervised component is calculated as follows:
\begin{equation}
    \mathcal{L}^{\text{sup}} \left( \mathbf{x}, \mathbf{y} \right) = \mathcal{L}^{\text{sup}}_{\text{cls}} \left( \mathbf{x}, \mathbf{y} \right) + \mathcal{L}^{\text{sup}}_{\text{centre}} \left( \mathbf{x}, \mathbf{y} \right) + \mathcal{L}^{\text{sup}}_{\text{box}} \left( \mathbf{x}, \mathbf{y} \right) + \mathcal{L}^{\text{sup}}_{\text{mask}} \left( \mathbf{x}, \mathbf{y} \right) + \mathcal{L}^{\text{sup}}_{\text{IoU}} \left( \mathbf{x}, \mathbf{y} \right),
    \label{eq:loss_sup}
\end{equation}
where $\mathcal{L}^{\text{sup}}_{\text{cls}}$ is the supervised classification loss, $\mathcal{L}^{\text{sup}}_{\text{centre}}$ is the supervised centreness loss, $\mathcal{L}^{\text{sup}}_{\text{box}}$ represents the supervised bounding box regression loss, and $\mathcal{L}^{\text{sup}}_{\text{mask}}$ is the supervised segmentation mask loss, and $\mathcal{L}^{\text{sup}}_{\text{mask\_IoU}}$ is the supervised segmentation mask scoring loss.
Regarding the pseudo-labelling loss, we use the following definition:
\begin{equation}
    \mathcal{L}^{\text{unsup}} \left( \mathbf{x}, \mathbf{\hat{y}} \right) = \mathbbm{1}_{\mathbf{\hat{y}_\text{cls}} > \tau_{\text{cls}}} \mathcal{L}^{\text{unsup}}_{\text{cls}} \left( \mathbf{x}, \mathbf{\hat{y}} \right) + \mathbbm{1}_{\mathbf{\hat{y}_\text{IoU}} > \tau_{\text{IoU}}} \mathcal{L}^{\text{unsup}}_{\text{mask}} \left( \mathbf{x}, \mathbf{\hat{y}} \right) + \mathcal{L}^{\text{unsup}}_{\text{IoU}} \left( \mathbf{x}, \mathbf{\hat{y}} \right),
    \label{eq:loss_unsup}
\end{equation}
where $\mathcal{L}^{\text{unsup}}_{\text{cls}}$ is the unsupervised classification loss, $\mathcal{L}^{\text{unsup}}_{\text{mask}}$ is the unsupervised segmentation mask loss, and $\mathcal{L}^{\text{unsup}}_{\text{IoU}}$ is the unsupervised segmentation mask scoring loss.
Regarding the particular loss functions implementation, we follow FCOS and CenterMask: 
$\mathcal{L}^{\text{sup}}_{\text{cls}}$ is focal loss \citep{lin2017focal}, 
$\mathcal{L}^{\text{sup}}_{\text{box}}$ is UnitBox IoU loss \citep{yu2016unitbox}, 
$\mathcal{L}^{\text{sup}}_{\text{centre}}$ is binary cross-entropy loss,
$\mathcal{L}^{\text{sup}}_{\text{mask}}$ is average binary cross-entropy loss \citep{he2017mask}, 
and $\mathcal{L}^{\text{sup}}_{\text{IoU}}$ is $L_2$ loss.
The same losses are used for unsupervised components (where applicable).

The student is trained using a standard stochastic gradient descent, whereas the teacher can be perceived as an ensemble of the students:
\begin{align}
    \theta_s &\leftarrow \theta_s - \gamma \frac{\partial \left(  \mathcal{L}^{\text{sup}} + \lambda \mathcal{L}^{\text{unsup}} \right) }{\partial \theta_s}, \label{eq:optim_student} \\
    \theta_t &\leftarrow \alpha \theta_t + (1 - \alpha) \theta_s.
    \label{eq:optim_teacher}
\end{align}
Here $\theta_t$ and $\theta_s$ represent the teacher and student model parameters respectively and $\alpha$ is the EMA coefficient (a hyperparameter).
Following \cite{liu2021unbiased}, the teacher trained in such a way is more robust to the sudden changes of decision boundaries caused by the minority classes in batches -- especially in the presence of pseudo-labels.
An important practical implication of this is the fact that there is no need to store gradients for the teacher model, which reduces GPU memory usage (compared to simply training two models).

%% file: figures/architecture.tikz
\tikzset{>=latex}
\begin{tikzpicture}

    \filldraw[fill=gray!20!white, draw=black, rounded corners] (0, 0) rectangle (.5, .5) node[pos=.5] {$x$};
    
    \filldraw[fill=red!20!white, draw=black, rounded corners] (1.5, 1) rectangle (4, 1.5) node[pos=.5] {$\texttt{Aug}_{\texttt{weak}}(x)$};
    \filldraw[fill=blue!20!white, draw=black, rounded corners] (1.5, -1) rectangle (4, -.5) node[pos=.5] {$\texttt{Aug}_{\texttt{strong}}(x)$};
   
    \coordinate (t_a) at (5,1);
    \coordinate (t_b) at (7,.75);
    \coordinate (t_c) at (7,1.75);
    \coordinate (t_d) at (5,1.5);
    \filldraw[fill=red!20!white, draw=black] (t_a) -- (t_b) -- (t_c) -- (t_d) -- (t_a) node[pos=.5, right] {Teacher $\theta_t$};

    \coordinate (s_a) at (5,-1);
    \coordinate (s_b) at (7,-1.25);
    \coordinate (s_c) at (7,-.25);
    \coordinate (s_d) at (5,-.5);
    \filldraw[fill=blue!20!white, draw=black] (s_a) -- (s_b) -- (s_c) -- (s_d) -- (s_a) node[pos=.5, right] {Student $\theta_s$};

    \filldraw[fill=red!20!white, draw=black, rounded corners] (8, 1) rectangle (9.5, 1.5) node[pos=.5] {$\hat{y}_{\texttt{pseudo}}$};
    \filldraw[fill=blue!20!white, draw=black, rounded corners] (8, -1) rectangle (9.5, -.5) node[pos=.5] {$\hat{y}$};

    \filldraw[fill=gray!20!white, draw=black, rounded corners] (10.5, -1) rectangle (12, -.5) node[pos=.5] {$\mathcal{L}_{\texttt{pseudo}}$};

    \coordinate (x_u) at (.25,.5);
    \coordinate (x_d) at (.25,0);
    
    \coordinate (augw_l) at (1.5,1.25);
    \coordinate (augw_r) at (4,1.25);
    
    \coordinate (augs_l) at (1.5,-.75);
    \coordinate (augs_r) at (4,-.75);

    \coordinate (t_l) at (5,1.25);
    \coordinate (t_r) at (7,1.25);
    \coordinate (s_l) at (5,-.75);
    \coordinate (s_r) at (7,-.75);
    \coordinate (s_r2) at (7,-.85);

    \coordinate (t_d) at (6, 0.86);
    \coordinate (s_u) at (6, -0.38);

    \coordinate (y_t) at (8,1.25);
    \coordinate (y_s) at (8,-.75);
    \coordinate (y_s2) at (8,-.85);

    \coordinate (y_tr) at (9.5,1.25);
    \coordinate (y_sr) at (9.5,-.75);
    \coordinate (y_sr2) at (9.5,-.85);
    \coordinate (l_u) at (11.25,-.5);
    \coordinate (l_l) at (10.5,-.75);
    \coordinate (l_l2) at (11.25,-1.);
    \coordinate (l_l3) at (11.25,-1.15);
    \coordinate (l_l4) at (7,-1.15);

    \draw[->] (x_u) -- (.25,1.25) -- (augw_l);
    \draw[->] (x_d) -- (.25,-.75) -- (augs_l);
    \draw[->] (augw_r) -- (t_l);
    \draw[->] (augs_r) -- (s_l);
    \draw[->] (t_r) -- (y_t);
    \draw[->] (s_r) -- (y_s);
    
    \draw[->] (y_tr) -- (11.25,1.25) -- (l_u);
    \draw[->] (y_sr) -- (l_l);

    \draw[gray, ->, dashed] (s_u) -- (t_d) node[right, pos=.5] {EMA};
    \draw[gray, ->, dashed] (l_l2) -- (l_l3) -- (l_l4) node[below, pos=.5] {Thresholded $\nabla$};

\end{tikzpicture}

%% file: 4_results.tex
\section{Evaluation}
\label{sec:4_results}

This section describes the evaluation of \ourMethodName.
We start with discussing the training setup, implementation details, and dataset.
Then, we present the result of our main experiment, which is followed by the detailed analysis and ablation studies of particular components of \ourMethodName.

\subsection{Setup}
\label{sec:4_results_setup}

\paragraph{Setup and implementation.}
All the experiments were performed either on \texttt{a2-highgpu-4g} instances on Google Cloud Platform (4xA100, 40 GB RAM each) or various machines with 8 GPUs (up to 16 GB RAM each) on the proprietary cluster (each of which contained Titan V, RTX 2080 Ti, or Titan X GPUs).
Polite Teacher was developed on CenterMask2 and Unbiased Teacher codebase -- both built on the Detectron2 framework \citep{wu2019detectron2}.

\paragraph{Data and evaluation metrics.}
We evaluate Polite Teacher on the MS-COCO 2017 dataset \citep{lin2014microsoft} using different supervision regimes (1\%, 2\%, 5\% and 10\% supervised).
The supervised-unsupervised split is taken from the Unbiased Teacher \cite{liu2021unbiased} -- while it was originally meant to be used for evaluation of semi-supervised object detection, it can be used with our method as well.
We report evaluation results on \texttt{val} subset, as the test one is not publicly available.
The reported metric used in this study is mask mAP (mean average precision, simply called AP later on), which is calculated as an average of AP with IoU thresholds set from 0.5 to 0.95 (with 0.05 intervals). 
Bounding box AP is also reported for selected experiments.

\paragraph{Hyperparameters.}
As base hyperparameters, we used the ones set in CenterMask2.
The EMA coefficient $\alpha$ for Teacher learning is set to $0.9996$.
The models were trained for up to 270,000 batches with stochastic gradient descent.
We used batch size 32 (16 supervised and 16 unsupervised samples) with a learning rate $\gamma = 0.006$, weight decay of 0.0001 and momentum 0.9.
Similarly to the CenterMask2, the learning rate has been decreased by a factor of 10 on steps 210,000 and 250,000.
However, such a long training was often not necessary, as models overfitted on much earlier stages.
These experiments have been early stopped.
Regarding the pseudo-label threshodling, we used $\tau_{\text{cls}}=0.6$ and $\tau_{\text{IoU}}=0.9$.
The unsupervised weight has been set to $\lambda=2$.
More details on the last three values are in Section \ref{sec:4_results_abl}.
We use ResNet-50 backbone \citep{he2016deep} for all the experiments.

\subsection{Results}

\begin{figure}
    \centering
    \begin{subfigure}{0.24\textwidth}
        \centering
        \includegraphics[width=\linewidth]{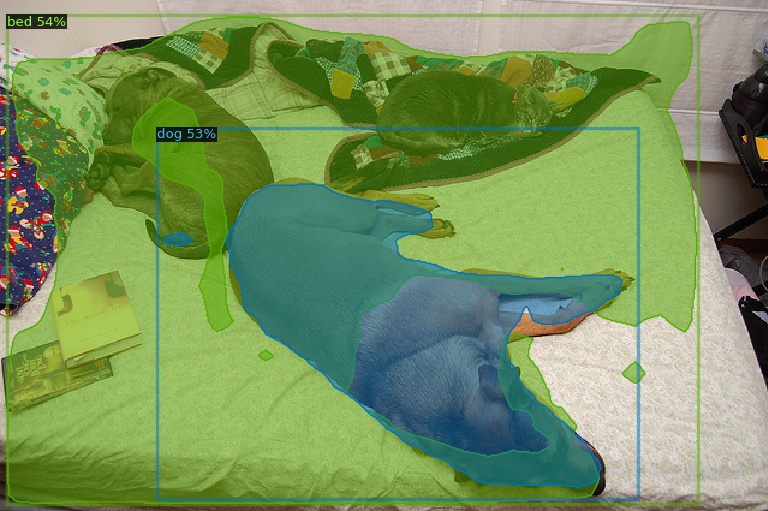}
      \end{subfigure}
      \begin{subfigure}{0.24\textwidth}
        \centering
        \includegraphics[width=\linewidth]{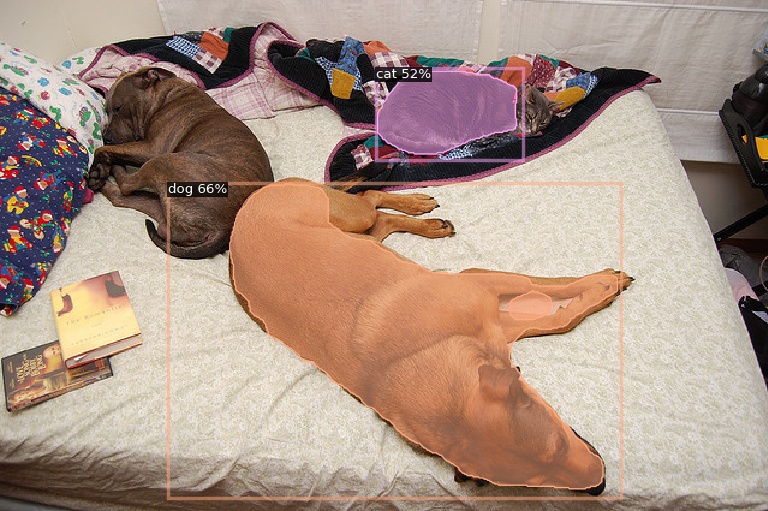}
      \end{subfigure}
      \begin{subfigure}{0.24\textwidth}
          \centering
          \includegraphics[width=\linewidth]{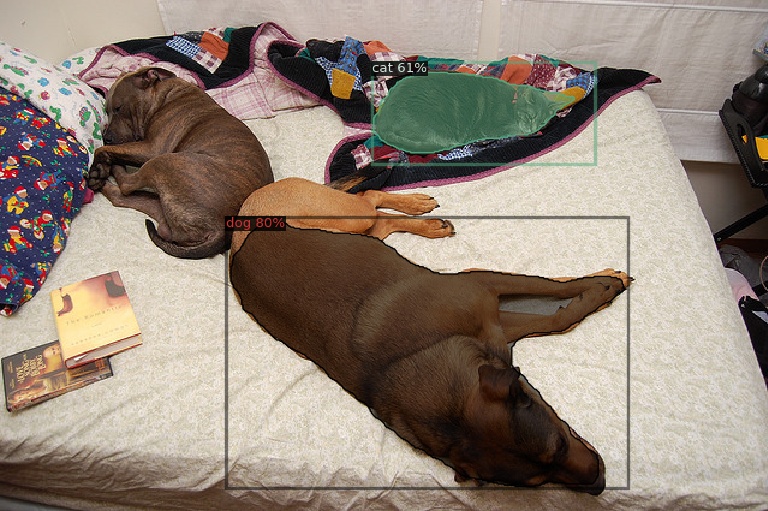}
      \end{subfigure}
      \begin{subfigure}{0.24\textwidth}
        \centering
        \includegraphics[width=\linewidth]{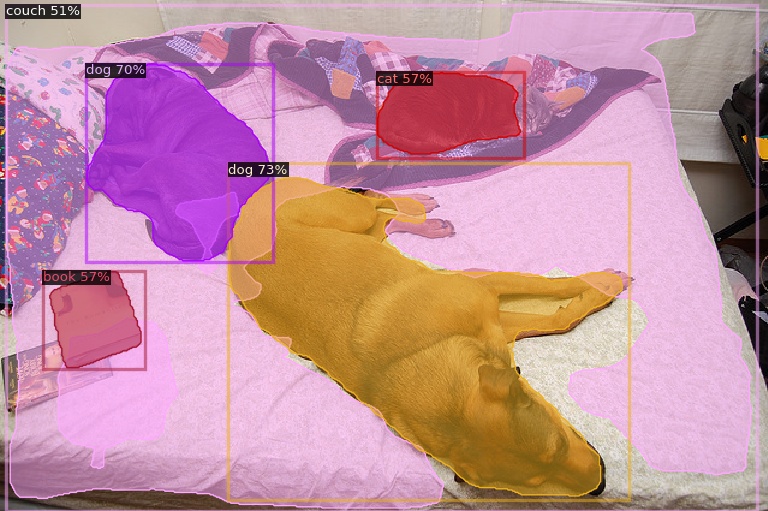}
      \end{subfigure}

      \begin{subfigure}{0.24\textwidth}
        \centering
        \includegraphics[width=\linewidth]{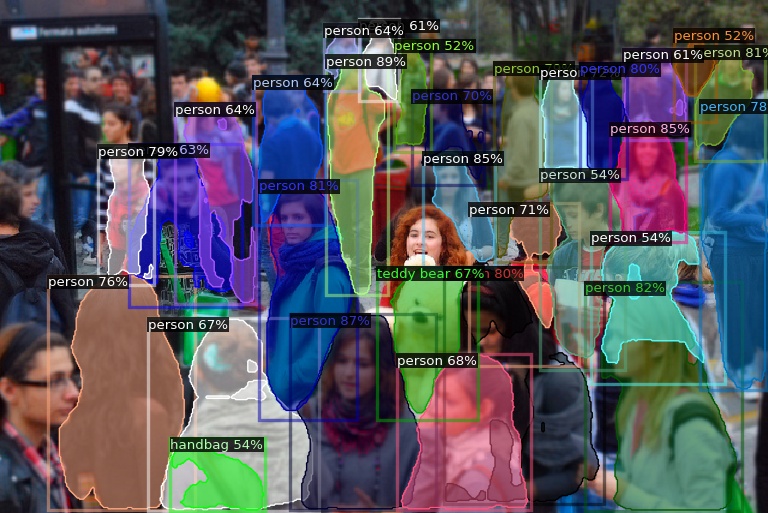}
      \end{subfigure}
      \begin{subfigure}{0.24\textwidth}
        \centering
        \includegraphics[width=\linewidth]{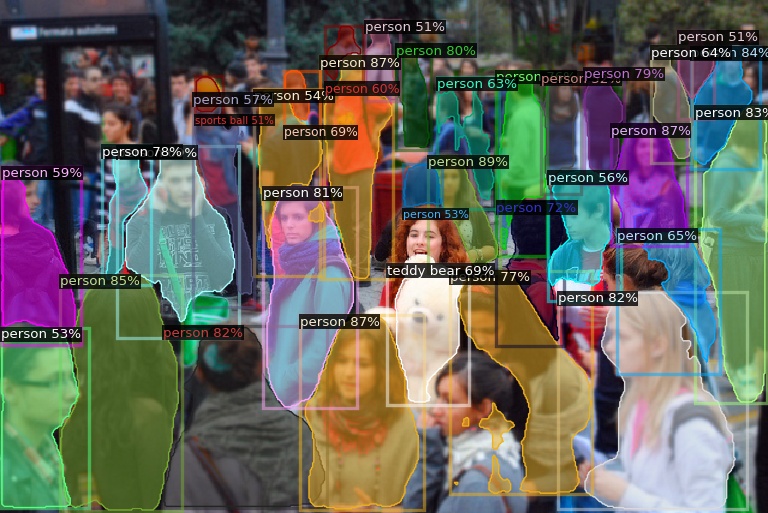}
      \end{subfigure}
      \begin{subfigure}{0.24\textwidth}
          \centering
          \includegraphics[width=\linewidth]{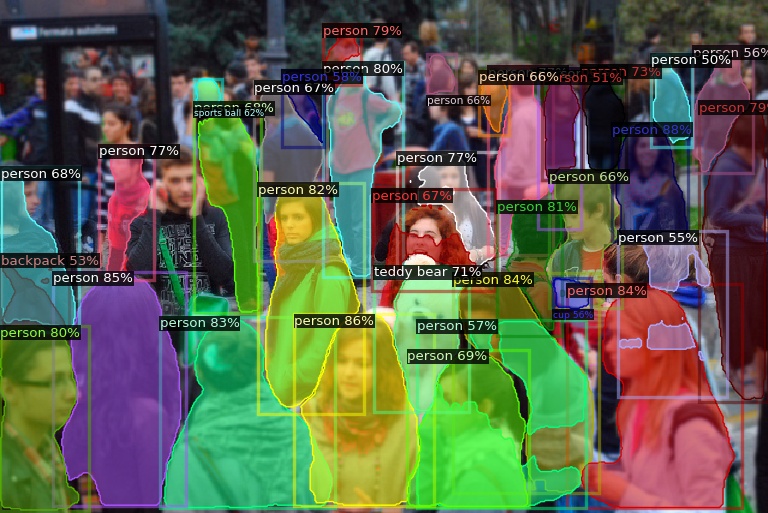}
      \end{subfigure}
      \begin{subfigure}{0.24\textwidth}
        \centering
        \includegraphics[width=\linewidth]{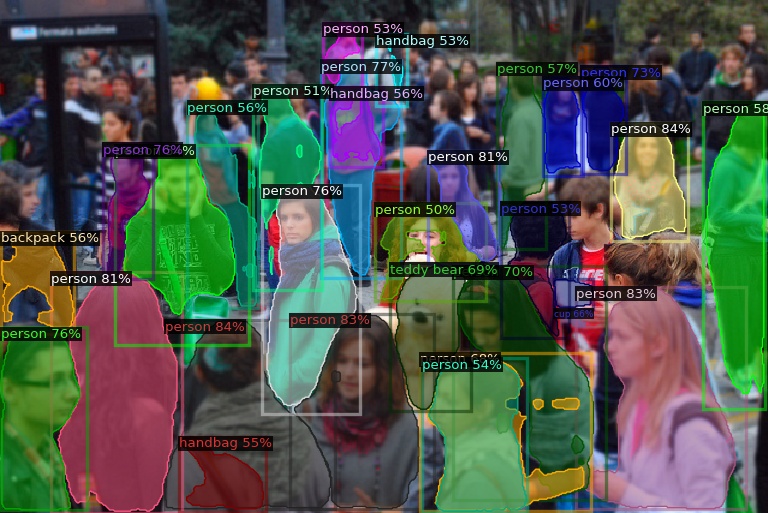}
      \end{subfigure}
 
      \begin{subfigure}{0.24\textwidth}
        \centering
        \includegraphics[width=\linewidth]{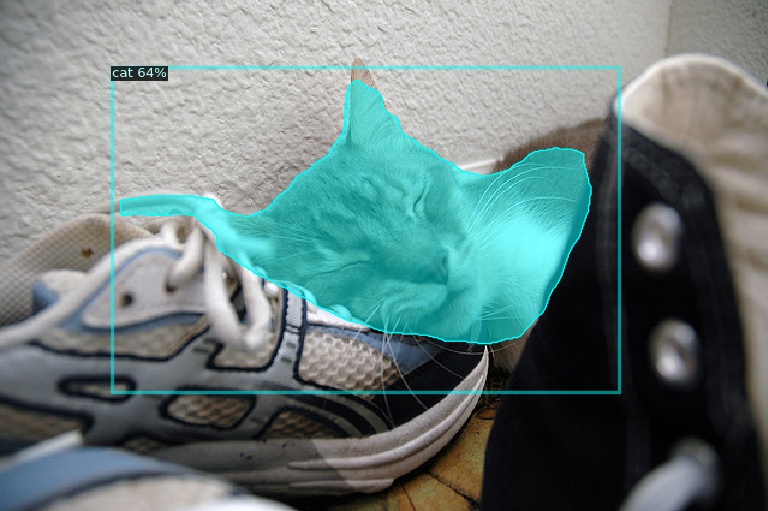}
      \end{subfigure}
      \begin{subfigure}{0.24\textwidth}
        \centering
        \includegraphics[width=\linewidth]{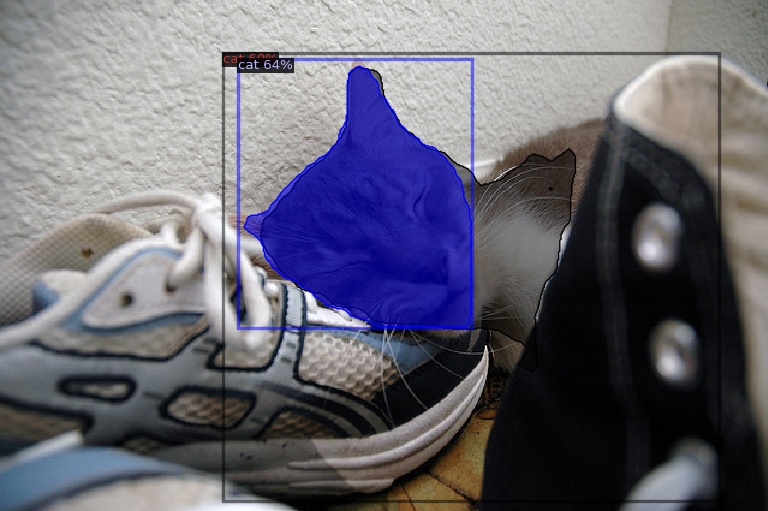}
      \end{subfigure}
      \begin{subfigure}{0.24\textwidth}
          \centering
          \includegraphics[width=\linewidth]{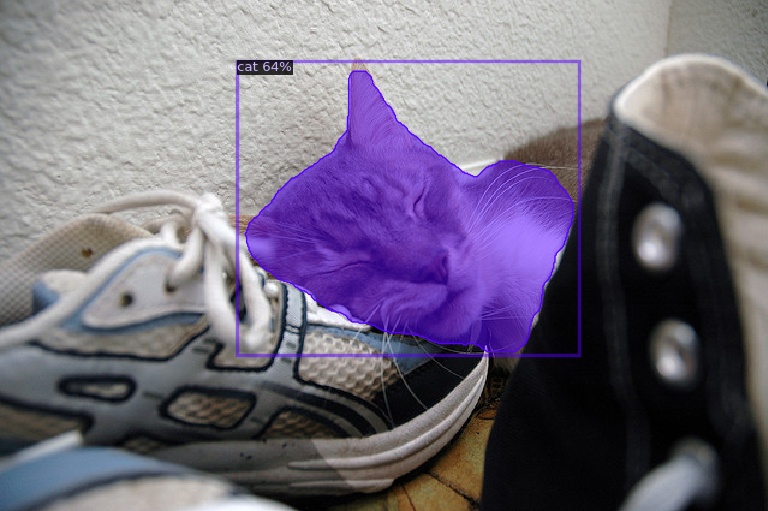}
      \end{subfigure}
      \begin{subfigure}{0.24\textwidth}
        \centering
        \includegraphics[width=\linewidth]{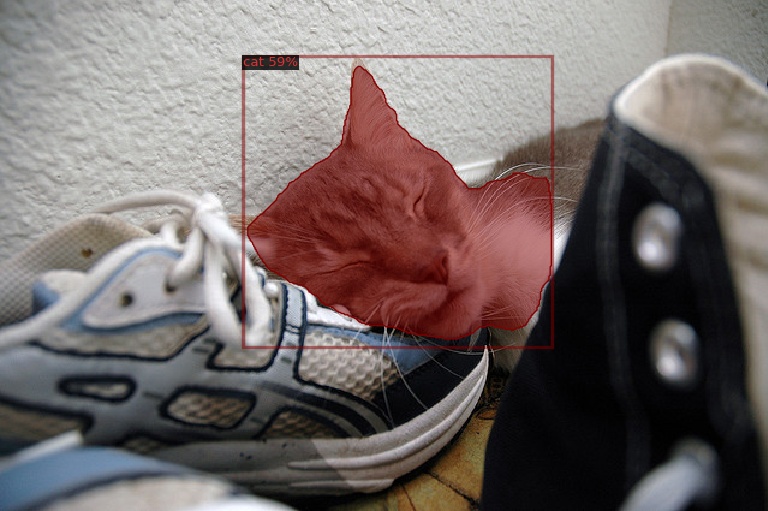}
      \end{subfigure}

      \begin{subfigure}{0.24\textwidth}
        \centering
        \includegraphics[width=\linewidth]{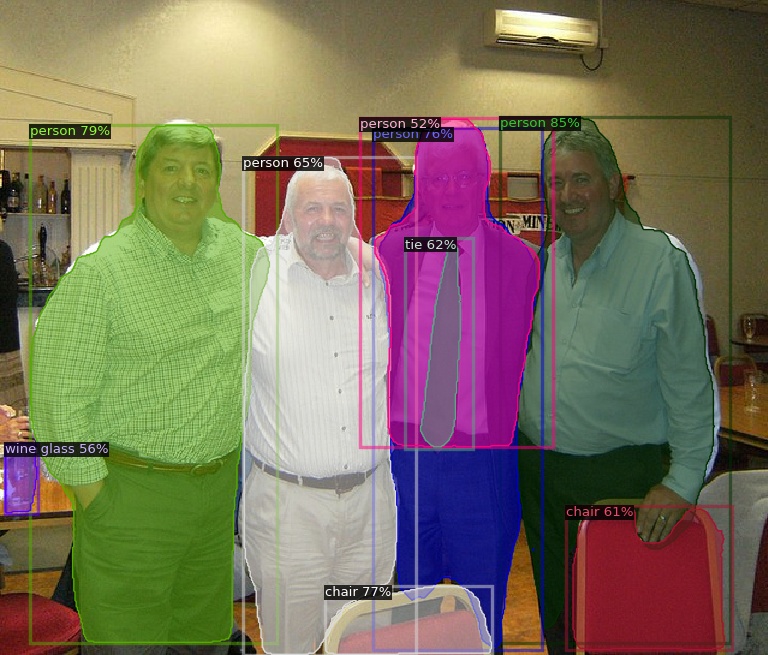}
        \caption{1\% supervised}
      \end{subfigure}
      \begin{subfigure}{0.24\textwidth}
        \centering
        \includegraphics[width=\linewidth]{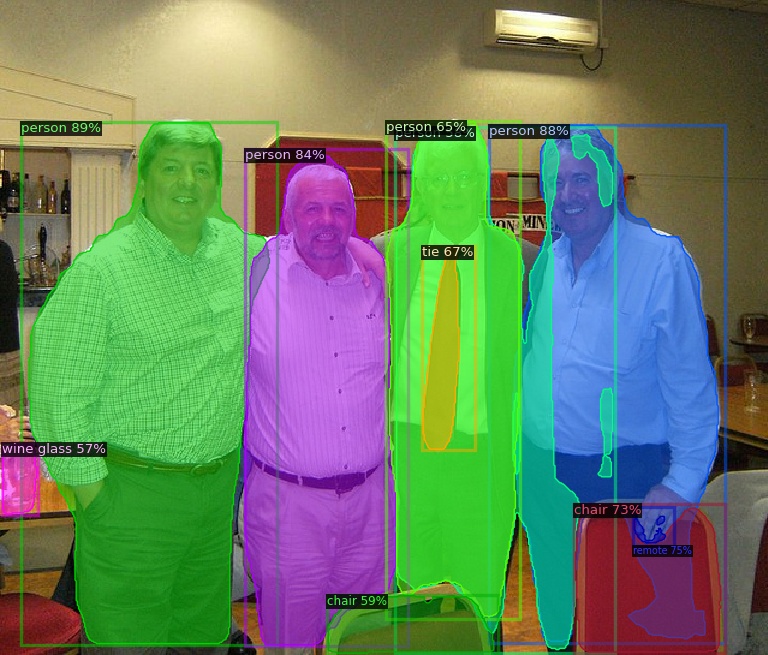}
        \caption{2\% supervised}
      \end{subfigure}
      \begin{subfigure}{0.24\textwidth}
          \centering
          \includegraphics[width=\linewidth]{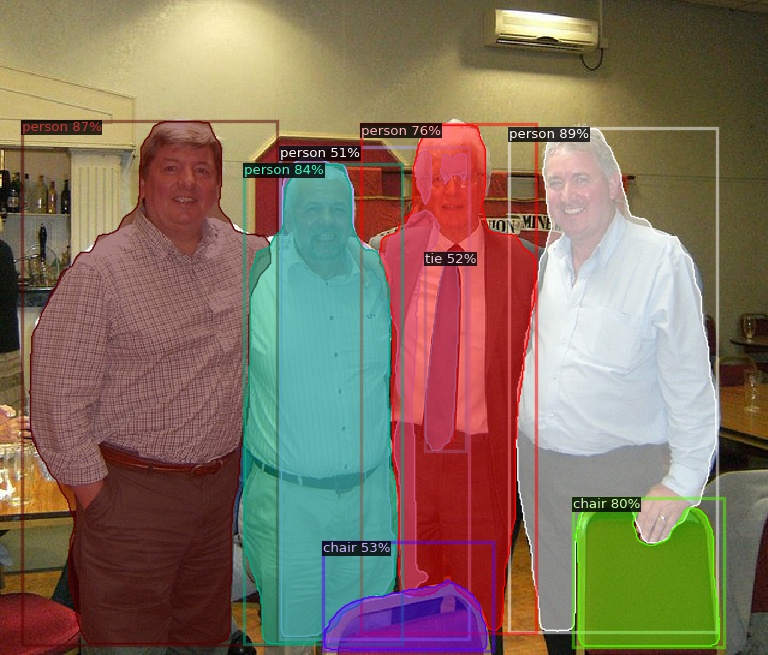}
          \caption{5\% supervised}
      \end{subfigure}
      \begin{subfigure}{0.24\textwidth}
        \centering
        \includegraphics[width=\linewidth]{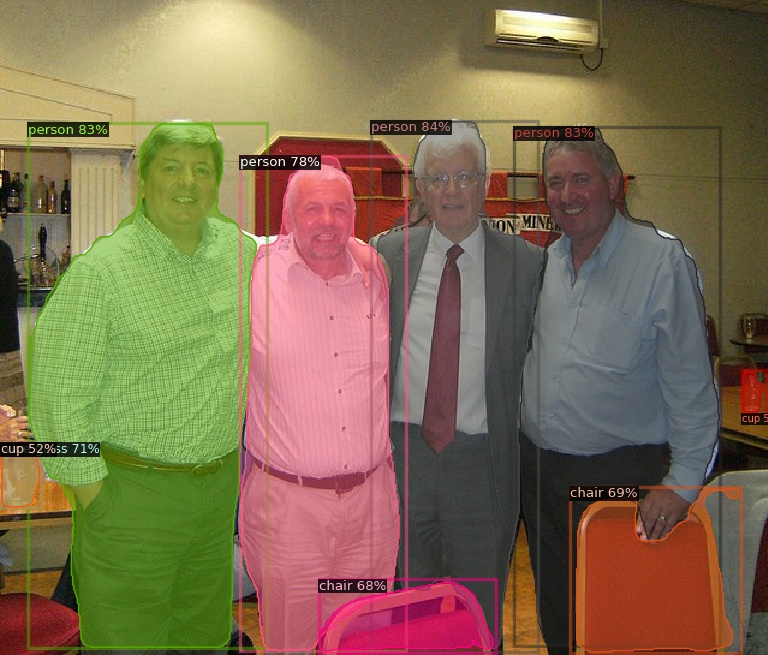}
        \caption{10\% supervised}
      \end{subfigure}      
    \caption{Qualitative \ourMethodName~results on COCO 2017 \texttt{val} with different supervision regimes.}
    \label{fig:qualitative_results}
  \end{figure}

Table \ref{tab:results} shows results for the main experiment conducted on MS-COCO 2017 validation dataset.
\ourMethodName~reached 18.33/22.28/26.46/30.08 mask AP on 1\%/2\%/5\%/10\% respectively, which stands for +8.26/+8.82/+8.42/+8.00 pp. change in this metric over the baseline CenterMask2 respectively.
Figure \ref{fig:qualitative_results} presents qualitative results from different models created in this experiment at different levels of supervision.

For the vast majority of our experiments, we thought that our method will be the first one devoted to semi-supervised instance segmentation.
The recent Noisy Boundaries (NB) approach \citep{wang2022noisy}, a concurrent work to \ourMethodName, is also concerned with this problem and has been evaluated on a similar percentage of supervision on COCO 2017.
However, these are different splits.
We did not perform direct comparisons, as we were not aware of this work for the majority of our research -- we report these results for scientific integrity, though.
In general, Noisy Boundaries reported a smaller increase in the mask AP (especially with low supervision), although for fair comparison such claims should be made after running the models on \emph{exactly} the same supervised/unsupervised data splits.
It is also unclear how much of this difference can be attributed to the different detectors (a two-staged Mask R-CNN has been used).
Following \cite{wang2022noisy}, we also report the results for Data Distillation (DD) method \cite{radosavovic2018data}, which was evaluated jointly with NB.
It was developed for the task of \emph{omni-supervised} (known also as \emph{webly-supervised}) learning, a special case of semi-supervised learning in which unlabelled data from the Internet are considered during the training.
At the heart of this approach lies the pipeline of different data transformations.
The results are later ensembled to provide pseudo-labels.

\begin{table}[]
    \centering
    \caption{Results with ResNet50 backbone. Oracle results reported by \cite{lee2020centermask}. The results for two-stage detectors are taken from the work of \cite{wang2022noisy}. Notice that it uses a random split of the dataset -- in particular, this is different from the one used by us. Therefore, these results cannot be directly compared, but we report them in order to trace the comparison with their baselines.}
    \label{tab:results}
    \begin{tabular}{lrrrr}
    \toprule
    \multicolumn{1}{c}{\multirow{2}{*}{Architecture}} & \multicolumn{4}{c}{\% supervised} \\ \cline{2-5} 
\multicolumn{1}{c}{} & \multicolumn{1}{c}{1} & \multicolumn{1}{c}{2} & \multicolumn{1}{c}{5} & \multicolumn{1}{c}{10} \\\midrule
\multicolumn{5}{c}{\emph{Mask AP, single-stage detectors (oracle: 34.70\%), COCO 2017 \texttt{val}, split from \cite{liu2021unbiased}}}\\\midrule
    CenterMask2 \citep{lee2020centermask}  & 10.07\phantom{$^{(+0.00)}$} & 13.46\phantom{$^{(+0.00)}$} & 18.04\phantom{$^{(+0.00)}$} & 22.08\phantom{$^{(+0.00)}$} \\ 
    \ourMethodName~(ours)                 & 18.33$^{(+8.26)}$  & 22.28$^{(+8.82)}$ & 26.46$^{(+8.42)}$ & 30.08$^{(+8.00)}$ \\\midrule
    \multicolumn{5}{c}{\emph{Mask AP, two-stage detectors (oracle: 34.50\%), COCO 2017 \texttt{val}, split from \cite{wang2022noisy}}}\\\midrule
    \multirow{1}{*}{Mask R-CNN \citep{he2017mask}}  & 3.5\phantom{0}\phantom{$^{(+0.00)}$} & 9.4\phantom{0}\phantom{$^{(+0.00)}$} & 17.3\phantom{0}\phantom{$^{(+0.00)}$} & 22.0\phantom{0}\phantom{$^{(+0.00)}$} \\ 
    \multirow{1}{*}{DD \citep{radosavovic2018data}} & 3.8\phantom{0}$^{(+0.30)}$ & 11.8\phantom{0}$^{(+2.40)}$          & 20.4\phantom{0}$^{(+3.10)}$           & 24.2\phantom{0}$^{(+2.20)}$ \\
    \multirow{1}{*}{NB \citep{wang2022noisy}}       & 7.7\phantom{0}$^{(+4.20)}$ & 16.3\phantom{0}$^{(+6.90)}$ & 24.9\phantom{0}$^{(+7.60)}$ & 29.2\phantom{0}$^{(+7.20)}$ \\ 
                                           \bottomrule
    \end{tabular}
\end{table}

\subsection{Detailed analysis and ablation studies}
\label{sec:4_results_abl}

In this section, a detailed analysis of the influence of hyperparameters and ablation studies is presented.
Unless otherwise specified, all the configuration is the same as in Section \ref{sec:4_results_setup}.
All experiments have been performed with the 5\% supervision regime.

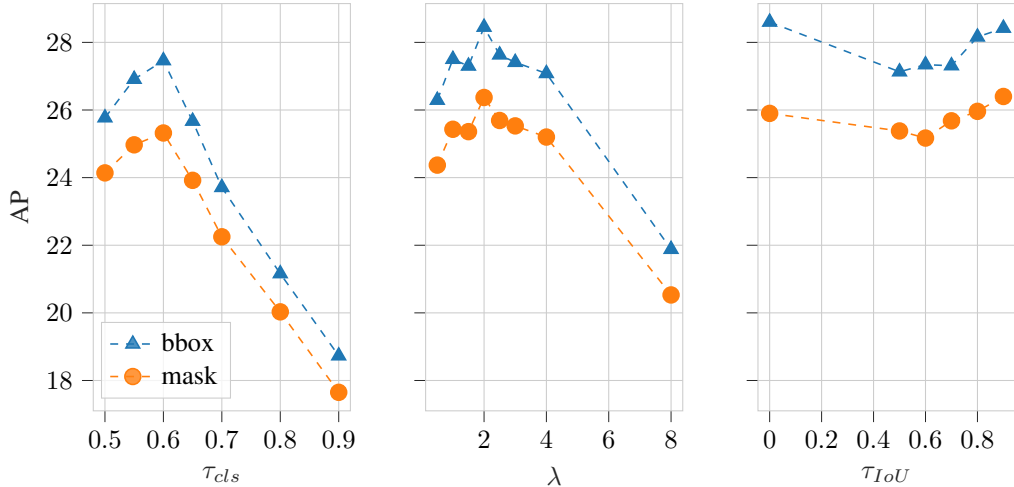
\begin{figure}
    \input{figures/ablations1-3.tex}
    \caption{Results for experiments controlled for different hyperparameters values in the 5\% supervision regime (see Section \ref{sec:4_results_abl}, as well as Table~\ref{tab:abl1}, \ref{tab:abl2}, and \ref{tab:abl3}).}
    \label{fig:abl1}
\end{figure}

\paragraph{Influence of bounding box filtering threshold.}
In this experiment, we investigate the importance of bounding box filtering thresholds.
To separate the influence of sole bounding box filtering, we did not include mask IoU in the optimisation -- it is a subject of another experiment.
That is, $\mathcal{L}^{\text{sup}}_{\text{IoU}} \left( \mathbf{x}_i, \mathbf{y}_i \right)$ and $\mathcal{L}^{\text{unsup}}_{\text{IoU}} \left( \mathbf{x}_j, \mathbf{\hat{y}}_j \right)$ has been not taken into account in equations \ref{eq:loss_sup} and \ref{eq:loss_unsup} respectively.
As it turns out, even this significantly improves mask AP over baselines.
Table \ref{tab:abl1} and Figure \ref{fig:abl1} (left) present AP values for this experiment.
The bounding box threshold value with the highest bounding box and mask AP was 0.6.
Interestingly, this is a slightly smaller threshold than in the original Unbiased Teacher paper (0.7).
The difference might stem from the different neural network architectures (Faster R-CNN vs CenterMask).
Note that this experiment used suboptimal $\lambda=0.75$ from Equation~\ref{eq:loss_main} and hence the results are slightly worse compared to the following experiments.
This is because the experiment to determine the correct unsupervised loss weight was yet to be carried out at this point. 

\begin{table}[]
    \centering
    \caption{Influence of bounding box filtering threshold (5\% supervision). }
    \label{tab:abl1}
    \begin{tabular}{lrrrrrrr}
    \toprule
    \multicolumn{1}{c}{\multirow{2}{*}{Metric}} & \multicolumn{7}{c}{$\tau_{\text{cls}}$} \\ \cline{2-8} 
\multicolumn{1}{c}{} & \multicolumn{1}{c}{0.5} & \multicolumn{1}{c}{0.55} & \multicolumn{1}{c}{0.6} & \multicolumn{1}{c}{0.65} & \multicolumn{1}{c}{0.7} & \multicolumn{1}{c}{0.8} & \multicolumn{1}{c}{0.9} \\\midrule
    AP (box) & 25.77 & 26.91 & 27.46 & 25.67 & 23.71 & 21.16 & 18.73\\
                                          AP (mask) & 24.14 & 24.97 & 25.32 & 23.92 & 22.25 & 20.03 & 17.65\\\bottomrule
    \end{tabular}
\end{table}

\paragraph{Influence of unsupervised loss weight.}
We also examine the influence of the weight of unsupervised loss, which is denoted as $\lambda$ in \eqref{eq:loss_main}.
Figure \ref{fig:abl1} (centre) and Table \ref{tab:abl2} presents detailed results of this study.
Similarly to the previous experiment, we did not include $\mathcal{L}^{\text{sup}}_{\text{IoU}} \left( \mathbf{x}_i, \mathbf{y}_i \right)$ and $\mathcal{L}^{\text{unsup}}_{\text{IoU}} \left( \mathbf{x}_j, \mathbf{\hat{y}}_j \right)$ from equations \ref{eq:loss_sup} and \ref{eq:loss_unsup} as the optimisation components.
We used $\tau_{\text{bbox}}=0.6$, which is the result of the previous experiment.
The highest mask AP has been obtained at $\lambda=2.0$.
Interestingly, in Unbiased Teacher, which is a similar architecture, this parameter was set to $\lambda=4.0$.

\begin{table}[]
    \centering
    \caption{Importance of unsupervised loss weight (5\% supervision). }
    \label{tab:abl2}
    \begin{tabular}{lrrrrrrrr}
    \toprule
    \multicolumn{1}{c}{\multirow{2}{*}{Metric}} & \multicolumn{8}{c}{$\lambda$} \\ \cline{2-9} 
\multicolumn{1}{c}{} & 0.5   & 1.0   & 1.5   & 2.0   & 2.5   &  3.0  &  4.0  &  8.0 \\\midrule
    AP (box)         & 26.29 & 27.50 & 27.30 & 28.45 & 27.63 & 27.41 & 27.08 & 21.88 \\
    AP (mask)        & 24.37 & 25.43 & 25.36 & 26.37 & 25.69 & 25.53 & 25.20 & 20.53 \\\bottomrule
    \end{tabular}
\end{table}

\paragraph{Influence of mask scoring filtering threshold.}
In this experiment, we investigate the importance of mask filtering threshold $\tau_{\text{IoU}}$.
We use $\tau_{\text{cls}}=0.6$ and $\lambda=2.0$, as these two values provided best results in the previous experiments.
Figure \ref{fig:abl1} (right) and Table \ref{tab:abl3} present the detailed results of this experiment.
Compared to the results without mask scoring, the best value ($\tau_{\text{IoU}}=0.9$) yielded insignificant differences in mask AP ($+0.03$ pp.) and bounding box AP ($-0.03$ pp.).
Interestingly, the conducted experiment displayed a convex-like U-shaped relationship between $\tau_{\text{cls}}$ and mask AP.
Passing all pseudo-masks resulted in the highest bbox AP, whereas filtering most of them yielded the highest mask AP.

\begin{table}[]
    \centering
    \caption{Influence of mask scoring filtering threshold.}
    \label{tab:abl3}
    \begin{tabular}{lrrrrrr}
    \toprule
    \multicolumn{1}{c}{\multirow{2}{*}{Metric}} & \multicolumn{6}{c}{$\tau_{\text{IoU}}$} \\ \cline{2-7} 
\multicolumn{1}{c}{} & \multicolumn{1}{c}{0.0} & \multicolumn{1}{c}{0.5} & \multicolumn{1}{c}{0.6} & \multicolumn{1}{c}{0.7} & \multicolumn{1}{c}{0.8} & \multicolumn{1}{c}{0.9} \\\midrule
    AP (box)  & 28.60 & 27.13 & 27.34 & 27.31 & 28.16 & 28.42 \\
    AP (mask) & 25.90 & 25.38 & 25.17 & 25.68 & 25.96 & 26.40 \\\bottomrule
    \end{tabular}
\end{table}

\paragraph{Ablation on pseudo-bounding box thresholding.}
For an ablation study, we compare the baseline CenterMask model to the Teacher-Student with bounding box thresholding.
Essentially, such a model is very similar to Unbiased Teacher \citep{liu2021unbiased}, which is proven to greatly improve results for semi-supervised object detection.
While raw CenterMask achieved 18.04\% on 5\% supervision, \ourMethodName~yielded 26.46\% mask AP, which is a +8.42 pp. increase (see tables \ref{tab:results} and \ref{tab:abl2}).
This suggests that much of the mask AP gain can be attributed to the Teacher-Student paradigm with bounding box thresholding.

\begin{figure}
    \input{figures/ablation6.tex}
    \caption{Mean results for \ourMethodName~training with 5\% supervised data, with (orange, 4 runs) and without (blue, 3 runs) mask scoring.}
    \label{fig:abl6}
\end{figure}
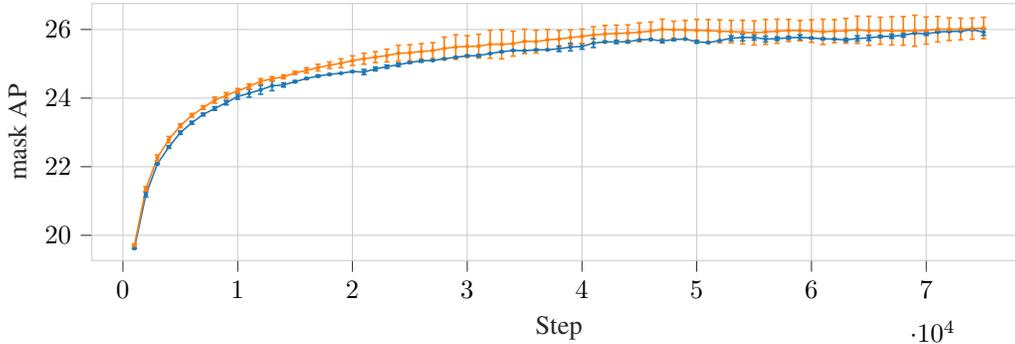

\paragraph{Ablation on pseudo-mask thresholding.}
In this ablation, we compare the model with bounding box thresholding to the model with bounding box and mask thresholding (that is, \ourMethodName).
Judging only by mask AP, the influence of the pseudo-mask filtering threshold on the final results can be easily neglected, as shown in Figure \ref{fig:abl1} (right).
However, applying mask scoring resulted in visibly faster convergence.
The model with mask scoring reached 26\% mask AP in 40k steps, whereas the model without it needed 74K steps to reach the same value, which is almost two times longer.
The highest mask AP values have been reached at 47k (26.39\%) and 99k step (26.37\%) respectively, which is also close to two times longer.
The detailed figures for this run are in tables \ref{tab:abl2} and \ref{tab:abl3}.
In order to check the stability of this behaviour, we repeat these experiments (Figure \ref{fig:abl6}).

\paragraph{Variance examination.}
Due to the computational limitations, we are not reporting results as a series of experiments with their means and standard deviations.
However, to assess the variance of the proposed method we carried out a separate experiment, in which we ran \ourMethodName~training with 5\% supervised data several times -- each time with a different seed value.
Figure \ref{fig:abl6} presents mean evaluation results per each step (batch), along with the standard deviations.
While the experiment has shown non-homogeneous variance for the variant with mask scoring, the maximum mask AP values are similar: 26.39, 26.39, 26.01, 25.85 with a mean of 26.16 ($\sigma=0.27$).
The variant without mask scoring achieved 26.03, 25.99, 25.99 max mask API (mean 26.00, $\sigma=0.02$) -- that is, an order of magnitude smaller variance, but at the expense of slower convergence and lower metrics value.
Interestingly, for the model with the mask scoring head, without the last run, the mean would be 26.26 ($\sigma=0.22$).
While it seems that the last run missed the local optima and much of the per-step variance can be attributed to it, we report all the obtained results for scientific integrity.
The high variance might suggest that another hyperparameter should be introduced (e.g. mask scoring head weight) or learning rate should be readjusted.

%% file: figures/ablations1-3.tex
\begin{tikzpicture}

\definecolor{darkorange25512714}{RGB}{255,127,14}
\definecolor{darkslategray38}{RGB}{38,38,38}
\definecolor{lightgray204}{RGB}{204,204,204}
\definecolor{steelblue31119180}{RGB}{31,119,180}

\begin{groupplot}[group style={group size=3 by 1}]
\nextgroupplot[
axis line style={lightgray204},
height=7cm,
legend cell align={left},
legend style={
  fill opacity=0.8,
  draw opacity=1,
  text opacity=1,
  at={(0.03,0.03)},
  anchor=south west,
  draw=lightgray204
},
tick align=outside,
tick pos=left,
width=5cm,
x grid style={lightgray204},
xlabel=\textcolor{darkslategray38}{\(\displaystyle \tau_{{cls}}\)},
xmajorgrids,
xmin=0.48, xmax=0.92,
xtick style={color=darkslategray38},
y grid style={lightgray204},
ylabel=\textcolor{darkslategray38}{AP},
ymajorgrids,
ymin=17.1025, ymax=29.1475,
ytick style={color=darkslategray38}
]
\addplot [semithick, steelblue31119180, dashed, mark=triangle*, mark size=3, mark options={solid}]
table {%
0.5 25.77
0.55 26.91
0.6 27.46
0.65 25.67
0.7 23.71
0.8 21.16
0.9 18.73
};
\addlegendentry{bbox}
\addplot [semithick, darkorange25512714, dashed, mark=*, mark size=3, mark options={solid}]
table {%
0.5 24.14
0.55 24.97
0.6 25.32
0.65 23.92
0.7 22.25
0.8 20.03
0.9 17.65
};
\addlegendentry{mask}

\nextgroupplot[
axis line style={lightgray204},
height=7cm,
scaled y ticks=manual:{}{\pgfmathparse{#1}},
tick align=outside,
tick pos=left,
width=5cm,
x grid style={lightgray204},
xlabel=\textcolor{darkslategray38}{\(\displaystyle \lambda\)},
xmajorgrids,
xmin=0.125, xmax=8.375,
xtick style={color=darkslategray38},
y grid style={lightgray204},
ymajorgrids,
ymin=17.1025, ymax=29.1475,
ytick style={color=darkslategray38},
yticklabels={}
]
\addplot [semithick, steelblue31119180, dashed, mark=triangle*, mark size=3, mark options={solid}]
table {%
0.5 26.29
1 27.5
1.5 27.3
2 28.45
2.5 27.63
3 27.41
4 27.08
8 21.88
};
\addplot [semithick, darkorange25512714, dashed, mark=*, mark size=3, mark options={solid}]
table {%
0.5 24.37
1 25.43
1.5 25.36
2 26.37
2.5 25.69
3 25.53
4 25.2
8 20.53
};

\nextgroupplot[
axis line style={lightgray204},
height=7cm,
scaled y ticks=manual:{}{\pgfmathparse{#1}},
tick align=outside,
tick pos=left,
width=5cm,
x grid style={lightgray204},
xlabel=\textcolor{darkslategray38}{\(\displaystyle \tau_{IoU}\)},
xmajorgrids,
xmin=-0.045, xmax=0.945,
xtick style={color=darkslategray38},
y grid style={lightgray204},
ymajorgrids,
ymin=17.1025, ymax=29.1475,
ytick style={color=darkslategray38},
yticklabels={}
]
\addplot [semithick, steelblue31119180, dashed, mark=triangle*, mark size=3, mark options={solid}]
table {%
0 28.6
0.5 27.13
0.6 27.34
0.7 27.31
0.8 28.16
0.9 28.42
};
\addplot [semithick, darkorange25512714, dashed, mark=*, mark size=3, mark options={solid}]
table {%
0 25.9
0.5 25.38
0.6 25.17
0.7 25.68
0.8 25.96
0.9 26.4
};
\end{groupplot}

\end{tikzpicture}

%% file: figures/ablation6.tex
\begin{tikzpicture}

\definecolor{darkorange25512714}{RGB}{255,127,14}
\definecolor{darkslategray38}{RGB}{38,38,38}
\definecolor{lightgray204}{RGB}{204,204,204}
\definecolor{steelblue31119180}{RGB}{31,119,180}

\begin{axis}[
axis line style={lightgray204},
height=5cm,
tick align=outside,
tick pos=left,
width=14cm,
x grid style={lightgray204},
xlabel=\textcolor{darkslategray38}{Step},
xmajorgrids,
xmin=-2701, xmax=78699,
xtick style={color=darkslategray38},
y grid style={lightgray204},
ylabel=\textcolor{darkslategray38}{mask AP},
ymajorgrids,
ymin=19.2655852671547, ymax=26.7557401502438,
ytick style={color=darkslategray38}
]
\path [draw=steelblue31119180, semithick]
(axis cs:999,19.6060468527497)
--(axis cs:999,19.63617370796);

\path [draw=steelblue31119180, semithick]
(axis cs:1999,21.1170023952838)
--(axis cs:1999,21.2553934380496);

\path [draw=steelblue31119180, semithick]
(axis cs:2999,22.0637806378588)
--(axis cs:2999,22.099274782063);

\path [draw=steelblue31119180, semithick]
(axis cs:3999,22.5356228001028)
--(axis cs:3999,22.6159096909454);

\path [draw=steelblue31119180, semithick]
(axis cs:4999,22.9418868167758)
--(axis cs:4999,23.0407995439013);

\path [draw=steelblue31119180, semithick]
(axis cs:5999,23.2359481834619)
--(axis cs:5999,23.3242210365088);

\path [draw=steelblue31119180, semithick]
(axis cs:6999,23.4856349992307)
--(axis cs:6999,23.5617791446496);

\path [draw=steelblue31119180, semithick]
(axis cs:7999,23.6452582592503)
--(axis cs:7999,23.7500448311631);

\path [draw=steelblue31119180, semithick]
(axis cs:8999,23.8029055626159)
--(axis cs:8999,23.9324803321595);

\path [draw=steelblue31119180, semithick]
(axis cs:9999,23.9686893315342)
--(axis cs:9999,24.1226294347419);

\path [draw=steelblue31119180, semithick]
(axis cs:10999,24.0271426644557)
--(axis cs:10999,24.2558570100235);

\path [draw=steelblue31119180, semithick]
(axis cs:11999,24.1126544566931)
--(axis cs:11999,24.3727708566207);

\path [draw=steelblue31119180, semithick]
(axis cs:12999,24.2164917308459)
--(axis cs:12999,24.5009536426893);

\path [draw=steelblue31119180, semithick]
(axis cs:13999,24.3227911063051)
--(axis cs:13999,24.4435319832945);

\path [draw=steelblue31119180, semithick]
(axis cs:14999,24.4495219240209)
--(axis cs:14999,24.5100575437526);

\path [draw=steelblue31119180, semithick]
(axis cs:15999,24.5488841802592)
--(axis cs:15999,24.5934765069967);

\path [draw=steelblue31119180, semithick]
(axis cs:16999,24.6147600310953)
--(axis cs:16999,24.6731262705812);

\path [draw=steelblue31119180, semithick]
(axis cs:17999,24.6762512845261)
--(axis cs:17999,24.707792091125);

\path [draw=steelblue31119180, semithick]
(axis cs:18999,24.7218376185702)
--(axis cs:18999,24.7289426141772);

\path [draw=steelblue31119180, semithick]
(axis cs:19999,24.763746745997)
--(axis cs:19999,24.7832579537101);

\path [draw=steelblue31119180, semithick]
(axis cs:20999,24.6837895398663)
--(axis cs:20999,24.83642581251);

\path [draw=steelblue31119180, semithick]
(axis cs:21999,24.792639884307)
--(axis cs:21999,24.908163236624);

\path [draw=steelblue31119180, semithick]
(axis cs:22999,24.8691836984206)
--(axis cs:22999,24.9640021332852);

\path [draw=steelblue31119180, semithick]
(axis cs:23999,24.9202313808371)
--(axis cs:23999,25.0229410739968);

\path [draw=steelblue31119180, semithick]
(axis cs:24999,25.0198225861172)
--(axis cs:24999,25.0557109628737);

\path [draw=steelblue31119180, semithick]
(axis cs:25999,25.0474507586272)
--(axis cs:25999,25.1260836346834);

\path [draw=steelblue31119180, semithick]
(axis cs:26999,25.0879276534535)
--(axis cs:26999,25.1043686925751);

\path [draw=steelblue31119180, semithick]
(axis cs:27999,25.1317838571068)
--(axis cs:27999,25.1615803498431);

\path [draw=steelblue31119180, semithick]
(axis cs:28999,25.1762845044125)
--(axis cs:28999,25.2035614326806);

\path [draw=steelblue31119180, semithick]
(axis cs:29999,25.2051541968904)
--(axis cs:29999,25.256186389454);

\path [draw=steelblue31119180, semithick]
(axis cs:30999,25.2041628082074)
--(axis cs:30999,25.2714752634886);

\path [draw=steelblue31119180, semithick]
(axis cs:31999,25.2989159650415)
--(axis cs:31999,25.2997638953914);

\path [draw=steelblue31119180, semithick]
(axis cs:32999,25.3354299014271)
--(axis cs:32999,25.3702611182669);

\path [draw=steelblue31119180, semithick]
(axis cs:33999,25.3759917252133)
--(axis cs:33999,25.4005750981103);

\path [draw=steelblue31119180, semithick]
(axis cs:34999,25.3773846742207)
--(axis cs:34999,25.3846570534812);

\path [draw=steelblue31119180, semithick]
(axis cs:35999,25.3872049729432)
--(axis cs:35999,25.427667768756);

\path [draw=steelblue31119180, semithick]
(axis cs:36999,25.3997741741724)
--(axis cs:36999,25.4189043638322);

\path [draw=steelblue31119180, semithick]
(axis cs:37999,25.3579270802197)
--(axis cs:37999,25.5180075206104);

\path [draw=steelblue31119180, semithick]
(axis cs:38999,25.3771741017819)
--(axis cs:38999,25.5962374582766);

\path [draw=steelblue31119180, semithick]
(axis cs:39999,25.4263885386321)
--(axis cs:39999,25.582857015974);

\path [draw=steelblue31119180, semithick]
(axis cs:40999,25.4740901672266)
--(axis cs:40999,25.7317178365169);

\path [draw=steelblue31119180, semithick]
(axis cs:41999,25.6140619621469)
--(axis cs:41999,25.662523934663);

\path [draw=steelblue31119180, semithick]
(axis cs:42999,25.5906314597125)
--(axis cs:42999,25.6858132932985);

\path [draw=steelblue31119180, semithick]
(axis cs:43999,25.609411756596)
--(axis cs:43999,25.6762925382445);

\path [draw=steelblue31119180, semithick]
(axis cs:44999,25.6857436715276)
--(axis cs:44999,25.7074969392308);

\path [draw=steelblue31119180, semithick]
(axis cs:45999,25.6913847844541)
--(axis cs:45999,25.7291143813988);

\path [draw=steelblue31119180, semithick]
(axis cs:46999,25.6094326146829)
--(axis cs:46999,25.7204013696921);

\path [draw=steelblue31119180, semithick]
(axis cs:47999,25.6752079107735)
--(axis cs:47999,25.7406042637057);

\path [draw=steelblue31119180, semithick]
(axis cs:48999,25.7118200915336)
--(axis cs:48999,25.7310471875191);

\path [draw=steelblue31119180, semithick]
(axis cs:49999,25.6018966242681)
--(axis cs:49999,25.6851237411291);

\path [draw=steelblue31119180, semithick]
(axis cs:50999,25.6071388183798)
--(axis cs:50999,25.6281873128051);

\path [draw=steelblue31119180, semithick]
(axis cs:51999,25.6677599927746)
--(axis cs:51999,25.682963273827);

\path [draw=steelblue31119180, semithick]
(axis cs:52999,25.6749750798117)
--(axis cs:52999,25.8182004267801);

\path [draw=steelblue31119180, semithick]
(axis cs:53999,25.6695425858251)
--(axis cs:53999,25.8669627954412);

\path [draw=steelblue31119180, semithick]
(axis cs:54999,25.6929195520478)
--(axis cs:54999,25.8374289078317);

\path [draw=steelblue31119180, semithick]
(axis cs:55999,25.6388864609109)
--(axis cs:55999,25.8071444737408);

\path [draw=steelblue31119180, semithick]
(axis cs:56999,25.6784710077248)
--(axis cs:56999,25.7856670232811);

\path [draw=steelblue31119180, semithick]
(axis cs:57999,25.7168114092878)
--(axis cs:57999,25.8010604474016);

\path [draw=steelblue31119180, semithick]
(axis cs:58999,25.6960372959222)
--(axis cs:58999,25.8461341823493);

\path [draw=steelblue31119180, semithick]
(axis cs:59999,25.7306152932433)
--(axis cs:59999,25.7735910462749);

\path [draw=steelblue31119180, semithick]
(axis cs:60999,25.7019109708181)
--(axis cs:60999,25.7543967582672);

\path [draw=steelblue31119180, semithick]
(axis cs:61999,25.7159998523477)
--(axis cs:61999,25.7253733369427);

\path [draw=steelblue31119180, semithick]
(axis cs:62999,25.6515624901255)
--(axis cs:62999,25.7401458838979);

\path [draw=steelblue31119180, semithick]
(axis cs:63999,25.6631367902904)
--(axis cs:63999,25.81743368497);

\path [draw=steelblue31119180, semithick]
(axis cs:64999,25.6791692076105)
--(axis cs:64999,25.8298233054104);

\path [draw=steelblue31119180, semithick]
(axis cs:65999,25.7674274531011)
--(axis cs:65999,25.8245779586828);

\path [draw=steelblue31119180, semithick]
(axis cs:66999,25.7431363921595)
--(axis cs:66999,25.8626769521602);

\path [draw=steelblue31119180, semithick]
(axis cs:67999,25.7758409038453)
--(axis cs:67999,25.8851049884887);

\path [draw=steelblue31119180, semithick]
(axis cs:68999,25.8921465531995)
--(axis cs:68999,25.8985511803617);

\path [draw=steelblue31119180, semithick]
(axis cs:69999,25.8282238424688)
--(axis cs:69999,25.9132833698521);

\path [draw=steelblue31119180, semithick]
(axis cs:70999,25.9055093009225)
--(axis cs:70999,25.9486370206921);

\path [draw=steelblue31119180, semithick]
(axis cs:71999,25.9267923387969)
--(axis cs:71999,25.9567385958548);

\path [draw=steelblue31119180, semithick]
(axis cs:72999,25.9463101600327)
--(axis cs:72999,25.9532653913182);

\path [draw=steelblue31119180, semithick]
(axis cs:73999,25.9823174655911)
--(axis cs:73999,26.0120456834161);

\path [draw=steelblue31119180, semithick]
(axis cs:74999,25.830242660459)
--(axis cs:74999,26.0128295949935);

\addplot [semithick, steelblue31119180, mark=-, mark size=1, mark options={solid}, only marks]
table {%
999 19.6060468527497
1999 21.1170023952838
2999 22.0637806378588
3999 22.5356228001028
4999 22.9418868167758
5999 23.2359481834619
6999 23.4856349992307
7999 23.6452582592503
8999 23.8029055626159
9999 23.9686893315342
10999 24.0271426644557
11999 24.1126544566931
12999 24.2164917308459
13999 24.3227911063051
14999 24.4495219240209
15999 24.5488841802592
16999 24.6147600310953
17999 24.6762512845261
18999 24.7218376185702
19999 24.763746745997
20999 24.6837895398663
21999 24.792639884307
22999 24.8691836984206
23999 24.9202313808371
24999 25.0198225861172
25999 25.0474507586272
26999 25.0879276534535
27999 25.1317838571068
28999 25.1762845044125
29999 25.2051541968904
30999 25.2041628082074
31999 25.2989159650415
32999 25.3354299014271
33999 25.3759917252133
34999 25.3773846742207
35999 25.3872049729432
36999 25.3997741741724
37999 25.3579270802197
38999 25.3771741017819
39999 25.4263885386321
40999 25.4740901672266
41999 25.6140619621469
42999 25.5906314597125
43999 25.609411756596
44999 25.6857436715276
45999 25.6913847844541
46999 25.6094326146829
47999 25.6752079107735
48999 25.7118200915336
49999 25.6018966242681
50999 25.6071388183798
51999 25.6677599927746
52999 25.6749750798117
53999 25.6695425858251
54999 25.6929195520478
55999 25.6388864609109
56999 25.6784710077248
57999 25.7168114092878
58999 25.6960372959222
59999 25.7306152932433
60999 25.7019109708181
61999 25.7159998523477
62999 25.6515624901255
63999 25.6631367902904
64999 25.6791692076105
65999 25.7674274531011
66999 25.7431363921595
67999 25.7758409038453
68999 25.8921465531995
69999 25.8282238424688
70999 25.9055093009225
71999 25.9267923387969
72999 25.9463101600327
73999 25.9823174655911
74999 25.830242660459
};
\addplot [semithick, steelblue31119180, mark=-, mark size=1, mark options={solid}, only marks]
table {%
999 19.63617370796
1999 21.2553934380496
2999 22.099274782063
3999 22.6159096909454
4999 23.0407995439013
5999 23.3242210365088
6999 23.5617791446496
7999 23.7500448311631
8999 23.9324803321595
9999 24.1226294347419
10999 24.2558570100235
11999 24.3727708566207
12999 24.5009536426893
13999 24.4435319832945
14999 24.5100575437526
15999 24.5934765069967
16999 24.6731262705812
17999 24.707792091125
18999 24.7289426141772
19999 24.7832579537101
20999 24.83642581251
21999 24.908163236624
22999 24.9640021332852
23999 25.0229410739968
24999 25.0557109628737
25999 25.1260836346834
26999 25.1043686925751
27999 25.1615803498431
28999 25.2035614326806
29999 25.256186389454
30999 25.2714752634886
31999 25.2997638953914
32999 25.3702611182669
33999 25.4005750981103
34999 25.3846570534812
35999 25.427667768756
36999 25.4189043638322
37999 25.5180075206104
38999 25.5962374582766
39999 25.582857015974
40999 25.7317178365169
41999 25.662523934663
42999 25.6858132932985
43999 25.6762925382445
44999 25.7074969392308
45999 25.7291143813988
46999 25.7204013696921
47999 25.7406042637057
48999 25.7310471875191
49999 25.6851237411291
50999 25.6281873128051
51999 25.682963273827
52999 25.8182004267801
53999 25.8669627954412
54999 25.8374289078317
55999 25.8071444737408
56999 25.7856670232811
57999 25.8010604474016
58999 25.8461341823493
59999 25.7735910462749
60999 25.7543967582672
61999 25.7253733369427
62999 25.7401458838979
63999 25.81743368497
64999 25.8298233054104
65999 25.8245779586828
66999 25.8626769521602
67999 25.8851049884887
68999 25.8985511803617
69999 25.9132833698521
70999 25.9486370206921
71999 25.9567385958548
72999 25.9532653913182
73999 26.0120456834161
74999 26.0128295949935
};
\path [draw=darkorange25512714, semithick]
(axis cs:999,19.6858676550498)
--(axis cs:999,19.7440992715249);

\path [draw=darkorange25512714, semithick]
(axis cs:1999,21.2919376076061)
--(axis cs:1999,21.4215910255116);

\path [draw=darkorange25512714, semithick]
(axis cs:2999,22.1769956159575)
--(axis cs:2999,22.3451544237154);

\path [draw=darkorange25512714, semithick]
(axis cs:3999,22.701517327391)
--(axis cs:3999,22.8815324467789);

\path [draw=darkorange25512714, semithick]
(axis cs:4999,23.1350198691601)
--(axis cs:4999,23.249568372417);

\path [draw=darkorange25512714, semithick]
(axis cs:5999,23.4433370375097)
--(axis cs:5999,23.5484527803004);

\path [draw=darkorange25512714, semithick]
(axis cs:6999,23.669986399775)
--(axis cs:6999,23.7772830399223);

\path [draw=darkorange25512714, semithick]
(axis cs:7999,23.8499908408552)
--(axis cs:7999,24.0294761696429);

\path [draw=darkorange25512714, semithick]
(axis cs:8999,23.9978090442443)
--(axis cs:8999,24.1601308666444);

\path [draw=darkorange25512714, semithick]
(axis cs:9999,24.1479825778595)
--(axis cs:9999,24.2780885891327);

\path [draw=darkorange25512714, semithick]
(axis cs:10999,24.2661612151323)
--(axis cs:10999,24.4149882675947);

\path [draw=darkorange25512714, semithick]
(axis cs:11999,24.4151138251076)
--(axis cs:11999,24.5644289071312);

\path [draw=darkorange25512714, semithick]
(axis cs:12999,24.500350327664)
--(axis cs:12999,24.6239162967867);

\path [draw=darkorange25512714, semithick]
(axis cs:13999,24.5663166176853)
--(axis cs:13999,24.6755123007717);

\path [draw=darkorange25512714, semithick]
(axis cs:14999,24.6887804862617)
--(axis cs:14999,24.776402771795);

\path [draw=darkorange25512714, semithick]
(axis cs:15999,24.7294406640037)
--(axis cs:15999,24.8841009390847);

\path [draw=darkorange25512714, semithick]
(axis cs:16999,24.7940668431894)
--(axis cs:16999,24.9857555063589);

\path [draw=darkorange25512714, semithick]
(axis cs:17999,24.8606680190316)
--(axis cs:17999,25.0545150482902);

\path [draw=darkorange25512714, semithick]
(axis cs:18999,24.8871073918294)
--(axis cs:18999,25.1461491389324);

\path [draw=darkorange25512714, semithick]
(axis cs:19999,24.9541769288388)
--(axis cs:19999,25.2308168150577);

\path [draw=darkorange25512714, semithick]
(axis cs:20999,24.9887728813003)
--(axis cs:20999,25.3071284172227);

\path [draw=darkorange25512714, semithick]
(axis cs:21999,25.0368166144585)
--(axis cs:21999,25.3531904045844);

\path [draw=darkorange25512714, semithick]
(axis cs:22999,25.0557265353164)
--(axis cs:22999,25.4248833584825);

\path [draw=darkorange25512714, semithick]
(axis cs:23999,25.0755809432575)
--(axis cs:23999,25.5382362717083);

\path [draw=darkorange25512714, semithick]
(axis cs:24999,25.0855050711278)
--(axis cs:24999,25.5574201913234);

\path [draw=darkorange25512714, semithick]
(axis cs:25999,25.1266265419846)
--(axis cs:25999,25.5871548148269);

\path [draw=darkorange25512714, semithick]
(axis cs:26999,25.1553507805945)
--(axis cs:26999,25.6148718832849);

\path [draw=darkorange25512714, semithick]
(axis cs:27999,25.1131276685165)
--(axis cs:27999,25.7789297503067);

\path [draw=darkorange25512714, semithick]
(axis cs:28999,25.1305939759616)
--(axis cs:28999,25.8450730238553);

\path [draw=darkorange25512714, semithick]
(axis cs:29999,25.1929444846062)
--(axis cs:29999,25.8213272515388);

\path [draw=darkorange25512714, semithick]
(axis cs:30999,25.1675602499839)
--(axis cs:30999,25.8606084282998);

\path [draw=darkorange25512714, semithick]
(axis cs:31999,25.1635604843779)
--(axis cs:31999,25.978031541379);

\path [draw=darkorange25512714, semithick]
(axis cs:32999,25.1457195968842)
--(axis cs:32999,25.9901017455841);

\path [draw=darkorange25512714, semithick]
(axis cs:33999,25.2249159788192)
--(axis cs:33999,25.9505000139176);

\path [draw=darkorange25512714, semithick]
(axis cs:34999,25.2981682707569)
--(axis cs:34999,26.010542781184);

\path [draw=darkorange25512714, semithick]
(axis cs:35999,25.3152517562876)
--(axis cs:35999,25.9783893341055);

\path [draw=darkorange25512714, semithick]
(axis cs:36999,25.3958727649454)
--(axis cs:36999,26.0007613369223);

\path [draw=darkorange25512714, semithick]
(axis cs:37999,25.4687843065338)
--(axis cs:37999,25.9747352857513);

\path [draw=darkorange25512714, semithick]
(axis cs:38999,25.5431239997388)
--(axis cs:38999,25.987233361101);

\path [draw=darkorange25512714, semithick]
(axis cs:39999,25.5895702656753)
--(axis cs:39999,26.0133702937119);

\path [draw=darkorange25512714, semithick]
(axis cs:40999,25.5966955889502)
--(axis cs:40999,26.0810578595361);

\path [draw=darkorange25512714, semithick]
(axis cs:41999,25.6189691611029)
--(axis cs:41999,26.1187093667291);

\path [draw=darkorange25512714, semithick]
(axis cs:42999,25.6397848112011)
--(axis cs:42999,26.1273298280812);

\path [draw=darkorange25512714, semithick]
(axis cs:43999,25.664170890745)
--(axis cs:43999,26.1338809433615);

\path [draw=darkorange25512714, semithick]
(axis cs:44999,25.6437240024427)
--(axis cs:44999,26.1923756221911);

\path [draw=darkorange25512714, semithick]
(axis cs:45999,25.6925720132392)
--(axis cs:45999,26.2179351889092);

\path [draw=darkorange25512714, semithick]
(axis cs:46999,25.7032985530414)
--(axis cs:46999,26.30189611096);

\path [draw=darkorange25512714, semithick]
(axis cs:47999,25.7401642931245)
--(axis cs:47999,26.2429337369658);

\path [draw=darkorange25512714, semithick]
(axis cs:48999,25.7115570818435)
--(axis cs:48999,26.2665456021774);

\path [draw=darkorange25512714, semithick]
(axis cs:49999,25.6541049015983)
--(axis cs:49999,26.2934315669076);

\path [draw=darkorange25512714, semithick]
(axis cs:50999,25.6462693190527)
--(axis cs:50999,26.2905840897608);

\path [draw=darkorange25512714, semithick]
(axis cs:51999,25.6106384096053)
--(axis cs:51999,26.2847664060685);

\path [draw=darkorange25512714, semithick]
(axis cs:52999,25.6084660933318)
--(axis cs:52999,26.2661028458772);

\path [draw=darkorange25512714, semithick]
(axis cs:53999,25.5771554137678)
--(axis cs:53999,26.2705847595721);

\path [draw=darkorange25512714, semithick]
(axis cs:54999,25.5498184781333)
--(axis cs:54999,26.2545333285074);

\path [draw=darkorange25512714, semithick]
(axis cs:55999,25.5652111306345)
--(axis cs:55999,26.2900439963186);

\path [draw=darkorange25512714, semithick]
(axis cs:56999,25.5981367782242)
--(axis cs:56999,26.3003503128402);

\path [draw=darkorange25512714, semithick]
(axis cs:57999,25.6496003462135)
--(axis cs:57999,26.2938191102684);

\path [draw=darkorange25512714, semithick]
(axis cs:58999,25.6721969529335)
--(axis cs:58999,26.2639412001427);

\path [draw=darkorange25512714, semithick]
(axis cs:59999,25.6253948900739)
--(axis cs:59999,26.2853993680437);

\path [draw=darkorange25512714, semithick]
(axis cs:60999,25.5632695050758)
--(axis cs:60999,26.2991982607323);

\path [draw=darkorange25512714, semithick]
(axis cs:61999,25.6182070401939)
--(axis cs:61999,26.2880896898476);

\path [draw=darkorange25512714, semithick]
(axis cs:62999,25.6000174425821)
--(axis cs:62999,26.322750196578);

\path [draw=darkorange25512714, semithick]
(axis cs:63999,25.5935077864192)
--(axis cs:63999,26.3896684449651);

\path [draw=darkorange25512714, semithick]
(axis cs:64999,25.5375446360143)
--(axis cs:64999,26.3851610143153);

\path [draw=darkorange25512714, semithick]
(axis cs:65999,25.5703935586973)
--(axis cs:65999,26.3570556676821);

\path [draw=darkorange25512714, semithick]
(axis cs:66999,25.54905947085)
--(axis cs:66999,26.373469749731);

\path [draw=darkorange25512714, semithick]
(axis cs:67999,25.5430647958922)
--(axis cs:67999,26.3764889608217);

\path [draw=darkorange25512714, semithick]
(axis cs:68999,25.5126579893062)
--(axis cs:68999,26.4152785646489);

\path [draw=darkorange25512714, semithick]
(axis cs:69999,25.574388982758)
--(axis cs:69999,26.3646264057308);

\path [draw=darkorange25512714, semithick]
(axis cs:70999,25.6364354279072)
--(axis cs:70999,26.3795319411724);

\path [draw=darkorange25512714, semithick]
(axis cs:71999,25.6798652776101)
--(axis cs:71999,26.33838232676);

\path [draw=darkorange25512714, semithick]
(axis cs:72999,25.6962221268414)
--(axis cs:72999,26.3301059600116);

\path [draw=darkorange25512714, semithick]
(axis cs:73999,25.7173632180648)
--(axis cs:73999,26.3330394232316);

\path [draw=darkorange25512714, semithick]
(axis cs:74999,25.728135956361)
--(axis cs:74999,26.3552552332142);

\addplot [semithick, darkorange25512714, mark=-, mark size=1, mark options={solid}, only marks]
table {%
999 19.6858676550498
1999 21.2919376076061
2999 22.1769956159575
3999 22.701517327391
4999 23.1350198691601
5999 23.4433370375097
6999 23.669986399775
7999 23.8499908408552
8999 23.9978090442443
9999 24.1479825778595
10999 24.2661612151323
11999 24.4151138251076
12999 24.500350327664
13999 24.5663166176853
14999 24.6887804862617
15999 24.7294406640037
16999 24.7940668431894
17999 24.8606680190316
18999 24.8871073918294
19999 24.9541769288388
20999 24.9887728813003
21999 25.0368166144585
22999 25.0557265353164
23999 25.0755809432575
24999 25.0855050711278
25999 25.1266265419846
26999 25.1553507805945
27999 25.1131276685165
28999 25.1305939759616
29999 25.1929444846062
30999 25.1675602499839
31999 25.1635604843779
32999 25.1457195968842
33999 25.2249159788192
34999 25.2981682707569
35999 25.3152517562876
36999 25.3958727649454
37999 25.4687843065338
38999 25.5431239997388
39999 25.5895702656753
40999 25.5966955889502
41999 25.6189691611029
42999 25.6397848112011
43999 25.664170890745
44999 25.6437240024427
45999 25.6925720132392
46999 25.7032985530414
47999 25.7401642931245
48999 25.7115570818435
49999 25.6541049015983
50999 25.6462693190527
51999 25.6106384096053
52999 25.6084660933318
53999 25.5771554137678
54999 25.5498184781333
55999 25.5652111306345
56999 25.5981367782242
57999 25.6496003462135
58999 25.6721969529335
59999 25.6253948900739
60999 25.5632695050758
61999 25.6182070401939
62999 25.6000174425821
63999 25.5935077864192
64999 25.5375446360143
65999 25.5703935586973
66999 25.54905947085
67999 25.5430647958922
68999 25.5126579893062
69999 25.574388982758
70999 25.6364354279072
71999 25.6798652776101
72999 25.6962221268414
73999 25.7173632180648
74999 25.728135956361
};
\addplot [semithick, darkorange25512714, mark=-, mark size=1, mark options={solid}, only marks]
table {%
999 19.7440992715249
1999 21.4215910255116
2999 22.3451544237154
3999 22.8815324467789
4999 23.249568372417
5999 23.5484527803004
6999 23.7772830399223
7999 24.0294761696429
8999 24.1601308666444
9999 24.2780885891327
10999 24.4149882675947
11999 24.5644289071312
12999 24.6239162967867
13999 24.6755123007717
14999 24.776402771795
15999 24.8841009390847
16999 24.9857555063589
17999 25.0545150482902
18999 25.1461491389324
19999 25.2308168150577
20999 25.3071284172227
21999 25.3531904045844
22999 25.4248833584825
23999 25.5382362717083
24999 25.5574201913234
25999 25.5871548148269
26999 25.6148718832849
27999 25.7789297503067
28999 25.8450730238553
29999 25.8213272515388
30999 25.8606084282998
31999 25.978031541379
32999 25.9901017455841
33999 25.9505000139176
34999 26.010542781184
35999 25.9783893341055
36999 26.0007613369223
37999 25.9747352857513
38999 25.987233361101
39999 26.0133702937119
40999 26.0810578595361
41999 26.1187093667291
42999 26.1273298280812
43999 26.1338809433615
44999 26.1923756221911
45999 26.2179351889092
46999 26.30189611096
47999 26.2429337369658
48999 26.2665456021774
49999 26.2934315669076
50999 26.2905840897608
51999 26.2847664060685
52999 26.2661028458772
53999 26.2705847595721
54999 26.2545333285074
55999 26.2900439963186
56999 26.3003503128402
57999 26.2938191102684
58999 26.2639412001427
59999 26.2853993680437
60999 26.2991982607323
61999 26.2880896898476
62999 26.322750196578
63999 26.3896684449651
64999 26.3851610143153
65999 26.3570556676821
66999 26.373469749731
67999 26.3764889608217
68999 26.4152785646489
69999 26.3646264057308
70999 26.3795319411724
71999 26.33838232676
72999 26.3301059600116
73999 26.3330394232316
74999 26.3552552332142
};
\addplot [semithick, steelblue31119180, mark=*, mark size=0.5, mark options={solid}]
table {%
999 19.6211102803548
1999 21.1861979166667
2999 22.0815277099609
3999 22.5757662455241
4999 22.9913431803385
5999 23.2800846099854
6999 23.5237070719401
7999 23.6976515452067
8999 23.8676929473877
9999 24.045659383138
10999 24.1414998372396
11999 24.2427126566569
12999 24.3587226867676
13999 24.3831615447998
14999 24.4797897338867
15999 24.5711803436279
16999 24.6439431508382
17999 24.6920216878255
18999 24.7253901163737
19999 24.7735023498535
20999 24.7601076761882
21999 24.8504015604655
22999 24.9165929158529
23999 24.971586227417
24999 25.0377667744954
25999 25.0867671966553
26999 25.0961481730143
27999 25.1466821034749
28999 25.1899229685466
29999 25.2306702931722
30999 25.237819035848
31999 25.2993399302165
32999 25.352845509847
33999 25.3882834116618
34999 25.3810208638509
35999 25.4074363708496
36999 25.4093392690023
37999 25.437967300415
38999 25.4867057800293
39999 25.5046227773031
40999 25.6029040018717
41999 25.6382929484049
42999 25.6382223765055
43999 25.6428521474202
44999 25.6966203053792
45999 25.7102495829264
46999 25.6649169921875
47999 25.7079060872396
48999 25.7214336395264
49999 25.6435101826986
50999 25.6176630655924
51999 25.6753616333008
52999 25.7465877532959
53999 25.7682526906331
54999 25.7651742299398
55999 25.7230154673258
56999 25.7320690155029
57999 25.7589359283447
58999 25.7710857391357
59999 25.7521031697591
60999 25.7281538645426
61999 25.7206865946452
62999 25.6958541870117
63999 25.7402852376302
64999 25.7544962565104
65999 25.7960027058919
66999 25.8029066721598
67999 25.830472946167
68999 25.8953488667806
69999 25.8707536061605
70999 25.9270731608073
71999 25.9417654673258
72999 25.9497877756755
73999 25.9971815745036
74999 25.9215361277262
};
\addplot [semithick, darkorange25512714, mark=*, mark size=0.5, mark options={solid}]
table {%
999 19.7149834632874
1999 21.3567643165588
2999 22.2610750198364
3999 22.791524887085
4999 23.1922941207886
5999 23.495894908905
6999 23.7236347198486
7999 23.939733505249
8999 24.0789699554443
9999 24.2130355834961
10999 24.3405747413635
11999 24.4897713661194
12999 24.5621333122253
13999 24.6209144592285
14999 24.7325916290283
15999 24.8067708015442
16999 24.8899111747742
17999 24.9575915336609
18999 25.0166282653809
19999 25.0924968719482
20999 25.1479506492615
21999 25.1950035095215
22999 25.2403049468994
23999 25.3069086074829
24999 25.3214626312256
25999 25.3568906784058
26999 25.3851113319397
27999 25.4460287094116
28999 25.4878334999084
29999 25.5071358680725
30999 25.5140843391418
31999 25.5707960128784
32999 25.5679106712341
33999 25.5877079963684
34999 25.6543555259705
35999 25.6468205451965
36999 25.6983170509338
37999 25.7217597961426
38999 25.7651786804199
39999 25.8014702796936
40999 25.8388767242432
41999 25.868839263916
42999 25.8835573196411
43999 25.8990259170532
44999 25.9180498123169
45999 25.9552536010742
46999 26.0025973320007
47999 25.9915490150452
48999 25.9890513420105
49999 25.9737682342529
50999 25.9684267044067
51999 25.9477024078369
52999 25.9372844696045
53999 25.9238700866699
54999 25.9021759033203
55999 25.9276275634766
56999 25.9492435455322
57999 25.971709728241
58999 25.9680690765381
59999 25.9553971290588
60999 25.9312338829041
61999 25.9531483650208
62999 25.9613838195801
63999 25.9915881156921
64999 25.9613528251648
65999 25.9637246131897
66999 25.9612646102905
67999 25.9597768783569
68999 25.9639682769775
69999 25.9695076942444
70999 26.0079836845398
71999 26.0091238021851
72999 26.0131640434265
73999 26.0252013206482
74999 26.0416955947876
};
\end{axis}

\end{tikzpicture}

%% file: 5_summary.tex
\section{Summary}
\label{sec:5_summary}

We presented \ourMethodName, a simple and effective architecture for semi-supervised instance segmentation.
Tested with a CenterMask, a single-stage detector, our approach yielded approx. +8 pp. mask AP on different supervision regimes with COCO 2017, while it introduces only three hyperparameters to tune.
A certain limitation of this study is the lack of validation on other datasets.
Similarly, more single-stage detectors, as well as two-stage detectors can be taken into consideration.
For more robust evaluation results, several runs of the experiments to explore variance on different values of supervision might be carried out.
Therefore, future work should consider validating methods with more detectors, backbones and datasets.
Providing a direct comparison with Noisy Boundaries \citep{wang2022noisy} might be considered as well.
A natural next step would consider taking pseudo-bounding boxes and pseudo-centre-ness regression into account.

%% file: 6_contributions.tex
\subsubsection*{Author Contributions}
\textbf{Dominik Filipiak:} (50\% of the work) conceptualization, methodology, software, validation, formal analysis, investigation, data curation, writing -- original draft, writing -- review \& editing, visualization, supervision, project administration.
\textbf{Andrzej Zapała:} (20\% of the work) software, validation, formal analysis, investigation, data curation, writing -- review \& editing, visualization.
\textbf{Piotr Tempczyk:} (10\% of the work) writing -- review \& editing, supervision.
\textbf{Anna Fensel:} (5\% of the work) resources, writing -- review \& editing, funding acquisition.
\textbf{Marek Cygan:} (15\% of the work) conceptualization, methodology, resources, writing -- review \& editing, supervision, funding acquisition.

%% file: 7_acknowledgements.tex
\subsubsection*{Acknowledgments}

This research was co-funded by Interreg \"Osterreich-Bayern 2014-2020 programme project KI-Net: Bausteine f\"ur KI-basierte Optimierungen in der industriellen Fertigung (grant agreement: AB 292).
Some experiments were performed using the Entropy cluster at the Institute of Informatics, University of Warsaw, funded by NVIDIA, Intel, the Polish National Science Center grant UMO2017/26/E/ST6/00622 and ERC Starting Grant TOTAL.
Marek Cygan is cofinanced by National Centre for Research and Development as a part of EU supported Smart Growth Operational Programme 2014-2020 (POIR.01.01.01-00-0392/17-00).

%% file: 8_reproducibility_statement.tex
\subsubsection*{Reproducibility Statement}

Along with the paper, we provide the source code for running our experiments.
This allows recreating the results from our experiments (up to the randomness).
The implementation is built on the Detectron2 framework \citep{wu2019detectron2}.
Our code is based on the freely available implementations of CenterMask2\footnote{\url{https://github.com/youngwanLEE/centermask2}} and Unbiased Teacher\footnote{\url{https://github.com/facebookresearch/unbiased-teacher}}, which were starting points for this research.
After the publication, the code for \ourMethodName~will be released on GitHub as well.

%% file: 9_ethics_statement.tex
\subsubsection*{Ethics Statement}

To the best of our knowledge, this work does not possess any direct ethical issues or a negative social impact.